\documentclass{article}

\usepackage{arxiv}

\usepackage{booktabs}       
\usepackage[authoryear]{natbib}
\usepackage{doi}

\usepackage{rotating} 
\usepackage{xcolor} 
\usepackage{array}
\usepackage{wrapfig}
\usepackage{amsmath, amsthm, amsfonts}
\usepackage{thmtools}
\usepackage{graphicx}
\usepackage{mathtools}
\usepackage{hyperref}
\usepackage{url}
\usepackage{array}
\usepackage{multirow}
\usepackage{caption}
\usepackage{subcaption}
\usepackage{makecell}

\makeatletter

\usepackage{enumitem} 

\usepackage{epstopdf}

\declaretheoremstyle[
spaceabove=6pt, spacebelow=6pt,
headfont=\normalfont\bfseries,
notefont=\mdseries, notebraces={(}{)},
bodyfont=\normalfont,
postheadspace=0.6em,
headpunct=:
]{mystyle}
\declaretheorem[style=mystyle, name=Hypothesis, preheadhook={}]{hyp}

\usepackage{arydshln}
\theoremstyle{definition}

\setcitestyle{authoryear,open={(},close={)}}
\usepackage{doi}

\title{Quantitative Discourse Cohesion Analysis of Scientific Scholarly Texts using Multilayer Networks}

\author{{Vasudha Bhatnagar} \\
	Department of Computer Science\\
	University of Delhi\\
	New Delhi,, India 110007 \\
	\And
	{Swagata Duari}\thanks{Corresponding Author. Email: \texttt{swagata.duari@gmail.com}} \\
	Department of Computer Science\\
	University of Delhi\\
	New Delhi,, India 110007 \\
	\And
	{S.K. Gupta} \\
	Department of Computer Science and Engineering\\
	Indraprastha Institute of Information Technology\\
	New Delhi, India 110020 \\
}

\renewcommand{\shorttitle}{\textit{arXiv} Template}

\hypersetup{
pdftitle={Quantitative Discourse Cohesion Analysis}
}

\begin{document}
\maketitle

\begin{abstract}
Discourse cohesion facilitates text comprehension and helps the reader form a coherent narrative. In this study, we aim to computationally analyze the discourse cohesion in scientific scholarly texts using multilayer network representation and quantify the writing quality of the document. Exploiting the hierarchical structure of scientific scholarly texts, we design section-level and document-level metrics to assess the extent of lexical cohesion in text. We use a publicly available dataset along with a curated set of contrasting examples to validate the proposed metrics by comparing them against select indices computed using existing cohesion analysis tools. We observe that the proposed metrics correlate as expected with the existing cohesion indices.

We also present an analytical framework, \textit{CHIAA} (\underline{CH}eck \underline{I}t \underline{A}gain, \underline{A}uthor), to provide pointers to the author for potential improvements in the manuscript with the help of the section-level and document-level metrics. The proposed CHIAA framework furnishes a clear and precise prescription to the author for improving writing by localizing regions in text with cohesion gaps. We demonstrate the efficacy of CHIAA framework using succinct examples from cohesion-deficient text excerpts in the experimental dataset.
\end{abstract}

\keywords{Discourse Cohesion \and Cohesion Metrics \and Computational Discourse Analysis \and Multilayer Networks \and Writing Quality}

\section{Introduction}
\label{sec:introduction}
Scholarly articles play a vital role in disseminating research findings and advancing the state of the art in science and technology. It is the most prevalent way for researchers to communicate their findings to the academic community. Apart from significance of the problem and novelty of solution, the writing quality of a scholarly article is an important determinant of the recognition it receives from the research community. A well-written scholarly document has a specific subject focus, which serves as the theme of the document. There is a global topic governing the written discourse, with no major topic shift\footnote{Survey articles sometimes may have minor topic shifts.}. According to \citet{danes1974functional}, progression of ideas in a scholarly article maps to the dynamic view of the discourse process. Concepts progressively evolve in an article as the author develops the manuscript, and ideas are connected smoothly from one section to the one or more following it.

The quality of a written discourse is characterized by two properties - \textit{coherence} and \textit{cohesion}. Coherence and cohesion are related yet distinct properties of a discourse. \textit{Discourse coherence} is a cognitive property of a text that signifies how well the text is making sense to the reader as a unified whole \citep{storrer2002coherence,wang2014short}. Coherence is a subjective property, which is dependent on the reader's background knowledge and comprehension abilities. \textit{Cohesion} is a semantic property of the document that represents how well the discourse entities are knit together throughout the document \citep{halliday1976cohesion,wang2014short}. Cohesion is an objective property of the language \citep{graesser2004coh}, which is enforced by repeated reference and semantic connection of entities in the document.

\begin{figure}[!h]
    \centering
    \begin{subfigure}{\columnwidth}
    \centering
        \includegraphics[scale = .5]{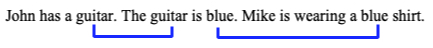}
        \caption{Text exhibit cohesion but no coherence.}
        \label{subfig:cohesion-no-coherence}
    \end{subfigure}
    \begin{subfigure}{\columnwidth}
    \centering
        \includegraphics[scale = .5]{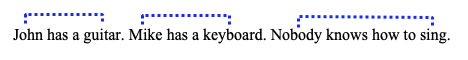}
        \caption{Text exhibit coherence but no cohesion.}
        \label{subfig:coherence-no-cohesion}
    \end{subfigure}
    \begin{subfigure}{\columnwidth}
    \centering
        \includegraphics[scale = .5]{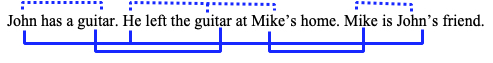}
        \caption{Text exhibit both coherence and cohesion.}
        \label{subfig:coherence-cohesion}
    \end{subfigure}
    \caption{Text snippets showing presence of coherence and cohesion.}
    \label{Fig:coherence-and-cohesion}
\end{figure}

We present three example snippets in Figure \ref{Fig:coherence-and-cohesion} to further establish the key difference between discourse coherence and cohesion.
In the example shown in Figure \ref{Fig:coherence-and-cohesion}, text snippet in Figure \ref{subfig:cohesion-no-coherence} seems to be a random set of sentences that make little to no sense to the reader as a unified whole. There is no semantic relationship between the contexts in which the entities appear in the sentences, e.g., the entity \textit{blue} is repeated in two unrelated contexts. Text snippet in Figure \ref{subfig:coherence-no-cohesion} makes slightly more sense than Figure \ref{subfig:cohesion-no-coherence} as there is an implicit relationship among the entities - for example, \textit{being able to sing} is semantically associated with \textit{possessing musical instruments}. In this case, the reader's prior knowledge fills the semantic gap. Text snippet in Figure \ref{subfig:coherence-cohesion} makes the most sense among the three examples as there is an explicit semantic relationship among the entities. The text is easily comprehensible as compared to the examples in Figures \ref{subfig:cohesion-no-coherence} and \ref{subfig:coherence-no-cohesion} due to the cohesion enforced by the repetition of entities that appear in similar contexts, e.g., the entity \textit{guitar} belonging to \textit{John} in two sentences. 

Discourse cohesion signifies the extent of semantic relationship among words or phrases in the text \citep{halliday1976cohesion} and is an important facilitator of text comprehension \citep{graesser2004coh, mcnamara2014automated}. Cohesion aids in the development of narrative structure, which strengthens the quality of writing. Cohesive cues present in the text aid the reader in forming a coherent narrative. Cohesion refers to linguistic features (such as lexical repetition and anaphora), which are explicitly realized in the surface structure of the text \citep{stubbs2001computer}. Lack of cohesion introduces difficulty in the process of comprehension, which is a result of not being able to connect entities and concepts correctly. The far-reaching significance of cohesion in terms of writing quality and comprehensibility motivates researchers to capture and quantify discourse cohesion in a written text. 

\citet{halliday1976cohesion} identify five situations that induce cohesion - reference, substitution, ellipsis, conjunction, and lexical cohesion\footnote{We provide examples of text excerpts to demonstrate the five situations that induce cohesion in Appendix \ref{app:example-cohesion}.}. The authors group the first four situations under \textit{grammatical cohesion}, which involves the syntactic structure of the document. Grammatical cohesion is enforced by using cohesive devices, which are commonly used words and phrases to connect ideas in different parts of the text. \textit{Lexical cohesion} involves the semantic relationship among words, which can be captured as recurrence of the words and their synonyms. The cohesion theory proposed by the group \citep{halliday1976cohesion, halliday1989language} is widely appreciated, expanded, and implemented in research \citep{levelt1989speaking, biber1991variation, foltz1998measurement, martin2001cohesion, graesser2004coh, mcnamara2010coh}. We congregate major ideas from earlier studies in this line to analyze scientific scholarly articles and quantitatively assess discourse cohesion. We hypothesize that a well-written text exhibits a high degree of cohesion. Accordingly, we focus on analyzing and quantifying the lexical cohesion in written discourse to assess the quality of writing. We do not analyze grammatical cohesion in our study as we focus on scientific scholarly texts written by researchers, and such texts are expected to maintain a certain standard of writing that is free of absurd and disastrous grammatical mistakes. 

We propose to model scientific scholarly documents as multilayer networks (MLN) to effectively capture the \textit{flow} of discourse. This model is a detailed map of the text that encapsulates the dynamics of evolution of writing of an article as the discourse progresses. The MLN model encodes the interactions between discourse entities within and across discourse units (sections, paragraphs, etc.). \textit{ To the best of the authors' knowledge, no existing work has used multilayer network representation of text for computational discourse analysis.}

\subsection{Research Objective and Contribution}
Cohesion gaps in text affect the readers' ability to form a coherent narrative as intended by the author. Currently available tools for cohesion analysis, such as Coh-metrix \citep{graesser2004coh, graesser2011coh} and TAACO \citep{crossley2016tool, crossley2019tool}, present the author with descriptive statistics (cohesion indices), which are automatically computed from linguistic features of the text. Coh-metrix provides 16 primary and $\approx 200$ additional indices of cohesion \citep{graesser2004coh}, and TAACO reports $\approx 150$ indices \citep{crossley2019tool}. However, these tools compute cohesion indices without providing any benchmark for comparison. In the absence of prescribed thresholds for guidance, users are compelled to define thresholds themselves, which can be any arbitrary value based on the users' judgment. Both Coh-metrix and TAACO tools do not provide any prescription for improvement by localizing regions with cohesion gaps. Furthermore, the cohesion indices derived from these tools are overwhelming in number and are not easy to interpret.

In this study, we design an analytical solution to identify and locate cohesion gaps in a scientific scholarly text. The objective of our study is to explore the following research questions: 
\begin{enumerate}
    \item \textit{Can we effectively exploit the document structure of scientific scholarly articles and map the flow of thematic progression to quantitatively assess discourse cohesion?}
    
    \item \textit{Does multilayer network representation of the scientific scholarly text empower computational discourse cohesion analysis by capturing the intricate relationships among discourse entities?}
    
    \item \textit{Can we design metrics to quantify the extent of discourse cohesion in texts by using apposite network properties distilled from the multilayer representation?}

    \item \textit{Can we appropriately locate cohesion gap in discourse by analyzing the semantic cues, and its lack thereof, based on the cohesion metric values?}
\end{enumerate}

We subscribe to the view that no writing is perfect and thus aim to objectively analyze scientific scholarly texts to identify gaps in writing - such as lack of cohesion and weak thematic progressions. Specifically, our contributions are as follows.
\begin{enumerate}
    \item We devise a novel representation of structured text documents based on multilayer complex networks. (Section \ref{sec:approach}).
    \item We translate cohesion of a scientific scholarly document to network-centric properties and design parameterless scoring functions to quantify the discourse cohesion at section and document level. (Sections \ref{sec:section-level}, and \ref{sec:document-level}).
    \item We curate a dataset of hundred (100) scientific scholarly documents published in predatory venues and analyze their writing quality to demonstrate the efficacy of the proposed approach against a public dataset of scholarly articles published in prestigious venues (Section \ref{sec:data}).
    \item We present empirical evidence supporting the proposed cohesion metrics and validate their efficacy using the experimental datasets (Section \ref{sec:experiments}).
    \item We propose \textit{CHIAA} framework to localize the lack of cohesion of text and weak thematic progression of concepts in the document. The framework identifies regions for potential improvements in text and reports them to the author for improvement (Section \ref{sec:author_suggestion}).
    \item We discuss in detail discussion on the theoretical and practical implication of this study, highlighting the advantages, potentials, and limitations of the proposed approach (Section \ref{sec:implications}).
    
\end{enumerate}

\section{Related Works}
\label{sec:related-works}
In this section, we review literature related to the following three areas - (i) Discourse Cohesion Analysis, (ii) Text Quality Assessment, and (iii) Multilayer network representation for Text Analysis - which are relevant to this study. 

\subsection{Discourse Cohesion Analysis}
\label{sec:related-works_CohAnalysis}
Early studies on cohesion analysis are primarily attributed to Halliday and Hasan, who consider cohesion as the marker for textual coherence \citep{halliday1976cohesion, halliday1989language}. They identify five situations that induce cohesion - reference, substitution, ellipsis, conjunction, and lexical cohesion - as discussed earlier \citep{halliday1976cohesion}. They perform extensive studies that transcend analysis of grammatical structure in discourse. \citet{carrell1982cohesion} criticizes the cohesion theory of \citeauthor{halliday1976cohesion} based on schema-theoretical views of text processing. \citeauthor{carrell1982cohesion} argues that text comprehension is an interactive process between the reader and their prior background knowledge, which \citeauthor{halliday1976cohesion} did not take into account in their work. Since modeling prior background knowledge of individual readers is an insurmountable task from a computational viewpoint, this line of thought is difficult to carry forward in the field of computational discourse analysis. 

\citet{graesser2004coh} propose a computational linguistic tool, \textit{Coh-Metrix}, to analyze discourse cohesion in text, which was later extended by \citet{mcnamara2010coh}. Coh-Metrix measures text cohesion, readability, and text difficulty on a range of word, sentence, paragraph, and discourse dimensions. Coh-metrix employs various NLP tools, like syntactic parser, part-of-speech tagger, lexicons, etc., to perform a wide range of linguistic and discourse analyses of the text document. Various extensions with incremental improvements to the original design of Coh-Metrix have been proposed so far \citep{graesser2011coh, graesser2014coh, dowell2016language}.

\citet{crossley2016tool} propose TAACO - a Tool for Automatic Analysis of Cohesion - which aids the user in assessing \textit{local} (sentence-level), \textit{global} (paragraph-level), and \textit{text} (document-level) cohesion. This study extends previous works on cohesion analysis \citep{graesser2004coh,mcnamara2010coh,graesser2011coh,graesser2014coh}, and presents multiple cohesion indices based on lexical overlapping, semantic overlapping, type-token ratio, connectives, etc. \citet{crossley2019tool} introduce TAACO 2.0, which is an expansion of TAACO with more cohesion indices. 

A contemporary study by \citet{jia2018modeling} captures discourse cohesion by modeling long-span dependencies of discourse units using memory networks, where the end goal is to perform discourse parsing using discourse cohesion. \citet{crossley2020linguistic} study linguistic features in texts and how they affect the writing quality. The author notes that discourse cohesion markers in text affect writing and development, and well-written essays demonstrate sophisticated lexical items, complex syntactic features, and greater cohesion. \citet{crossley2022linguistic} extend this assertion and extensively examine the writing quality and writing development from a linguistic perspective.

\subsection{Text Quality Assessment}
\label{sec:related-works_textQuality}
\citet{louis2013makes} show that distinguishing articles based on writing quality is fairly accurate in the domain of scientific journalism. The authors consider several textual features that capture the readability and well-written nature of the text. This work sets the pretext for measuring overall (global) article quality. \citet{amancio2015comparing} bring to fore complex network features that distinguish artificially generated papers from real manuscripts. Though the objective of this study is text classification, their analysis shows that both classes of text, when represented as graphs, display distinct topological properties. 

Several works in text quality assessment focus on Wikipedia articles, varying on tasks like binary classification of featured vs. not featured articles \citep{lipka2010identifying, arnold2016network} and multi-class classification of all Wikipedia quality classes \citep{dang2017end}. \citet{arnold2016network} analyze the text quality in German Wikipedia articles (featured vs. non-featured) and argue that network motifs are potentially impactful in improving text quality assessment. 

\citet{amancio2015comparing} analyzes two categories of texts using graph representation, where the author considers two extreme types of text, i.e., real vs. artificially generated. Our study on scholarly texts for assessing discourse coherence and cohesion using complex networks is influenced by the author's observation that graph-theoretic properties are capable of distinguishing the two categories of texts. It is noteworthy that artificially generated texts fail to adequately capture the intricacies of real-life writing quality as exposed by the wide range of semantic and cognitive cues originating from the author's expertise and background knowledge. We, therefore, delve into an analysis of articles published in two different types of venues - prestigious venues with a rigorous peer-review process and predatory venues with relatively relaxed writing criteria.

\subsection{Multilayer Networks for Text Analysis}
\label{sec:related-works_MLN}
Multilayer networks conveniently map inherently complex systems, like multiple types of relations in social networks, multiple modes of transport in transportation networks, and multiple relations among biological entities in biological networks \citep{boccaletti2014structure}. However, despite their overwhelming applicability, multilayer networks have been relatively under-explored for text analysis tasks. Recent works on multi-document text summarization \citep{tohalino2018extractive} and document analysis \citep{sebestyen2020multilayer, hyland2021multilayer} use multilayer network representations. 

\citet{tohalino2018extractive} model each document in the corpus as a layer in the network, where the sentences in the document constitute the set of vertices in the layer. Topological properties like degree, strength, PageRank, accessibility, shortest paths, etc., are extracted from the network. A multi-document summary is generated by selecting the best-ranked sentences based on these properties. \citet{sebestyen2020multilayer} present a method called MUNCoDA to analyze inconsistent textual information in documents from the same subject area. The method models each document as a layer in the network with a fixed set of words as nodes. The word pairs within a layer are linked by a weighted edge denoting the frequency of the words' connection in the given document. 

\citet{hyland2021multilayer} explore multilayer networks to investigate clustering and infer latent relationships from collections of documents. The authors model the document collection as a multilayer network with the documents as nodes in each layer. The edges connecting these nodes represent a different relationship (view) at each layer. Posing as a generalized topic modeling method, the authors note that the multilayer network approach succeeds in integrating multiple data types, which creates more nuanced communities of documents and improves the ability to predict missing links.

\textit{Recognizing the potential of multilayer networks for text analytics applications, we explore multilayer networks for effectively modeling and analyzing scientific scholarly texts for discourse cohesion analysis. To the best of the authors' knowledge, no existing work has used multilayer network representation to analyze textual discourse. We distinguish our work from the existing cohesion analysis tools by presenting the user with easy-to-understand and intuitive metrics along with pointers to the regions in text where they could potentially improve the writing.}
\section{Our Approach}
\label{sec:approach}
Scholarly documents are usually long and comprise topically coherent text segments (sections), each containing a number of text passages such as subsections, paragraphs, and sentences. \citet{salton1996automatic} also assert such logically structured hierarchy of discourse units (i.e., sections, subsections, paragraphs, etc.) in text documents. A well-written text displays cohesion and consistency (with respect to context) at each hierarchical level and also at a global level \citep{halliday1989language}. In this study, we divide the text into three hierarchical levels: the whole document, its constituent sections, and the sentences corresponding to each section. We model the text as a multilayer network and device network-centric measurements that adequately capture cohesion in the text. The multilayer network representation aids in localizing potentially weak areas in the text to provide suggestions to the author to improve the writing in those areas. 

We perform a two-level cohesion analysis in scientific scholarly documents, as described below.
\begin{enumerate}[nolistsep]
    \item \textbf{Cohesion at section level:} The section-level analysis is based on the premise that cohesive texts exhibit a strong co-occurrence relationship among words and phrases present in the text segment. We represent the section text as a complex network and model the discourse entities as nodes and their relationships as edges.  We analyze the consistency of ideas within individual sections of the document by finding cohesive cues within discourse entities of the section (Section \ref{sec:section-level}).
    \item \textbf{Cohesion at document level:} We analyze the global cohesion of a document by taking cognizance of the insights from section-level analysis and quantify the interplay of discourse entities across sections of the document. Using the section-level networks as individual layers, we model a multilayer network (MLN) representation of the document that encodes the relation between discourse entities across sections of the text (Section \ref{sec:document-level}). 
\end{enumerate}

We present our work as a proof of concept, where we validate the effectiveness of the network properties in assessing writing quality in terms of lexical cohesion in the text. To the best of the authors' knowledge, no prior work has addressed a similar objective, i.e., to quantify writing quality by using concise and precise metrics to gauge the extent of lexical cohesion in text. Due to this reason, we are unable to perform an empirical comparison with relevant studies. However, we provide empirical evidence, wherever appropriate, which establish agreement between the proposed metrics and those computed by an existing linguistic tool (TAACO \citep{crossley2016tool,crossley2019tool}). It is noteworthy that we do not restrict our analysis to only long documents that can be segmented into logically connected units (e.g., sections, chapters, etc.). If the document is short and can not be segmented further, we can perform section-level analysis to assess local cohesion and identify regions of weak cohesion, if any.

\section{Cohesion at Section Level}
\label{sec:section-level}
Scientific scholarly articles are written by experts and are targeted at a learned audience with sufficient familiarity with the subject. Scientific discourses often include specialized terms of the discipline woven together deftly at an advanced reading level. These specialized terms are referred to as keywords or key-entities, which are typically nouns\footnote{Discourse entities consist primarily of noun phrases \citep{barzilay2008modeling}. In the interest of simplicity, we consider only nouns as discourse entities.} \citep{barzilay2008modeling}. A cohesive discourse is expected to exhibit repeated use of key-entities either as synonyms, hyponyms, or hypernyms in similar and related contexts \citep{halliday1976cohesion, halliday1989language, morris1991lexical}. In other words, key-entities of a cohesive text maintain thematic consistency throughout the discourse.

\subsection{Modeling Text as Complex Network}
\label{subsec:text-as-cn}
In order to examine section-level cohesion of individual sections in scientific scholarly texts, we assess the extent of cohesion among the key-entities discussed in the section text. We model the section text as a \textit{complex network} that captures the interactions between constituting key-entities in the text. The entities are typically modeled as nodes or vertices in the network, and their interactions are modeled as links or edges. The links can be directed based on the direction of interaction and have weights based on the strength of interaction between the entities. Mathematically, a complex network is defined as a tuple $L = \left(V, E, W\right)$, where $V$ is the set of nodes (vertices), $E \subseteq V \times V$ is the set of links (edges), and $W$ is the matrix of size $|V| \times |V|$, where each element represent the edge weight for corresponding edge in $E$.

We identify key-entities from the section using a keyword extraction algorithm and retain them as the set of nodes $V$ in network $L$. Two key-entities are connected with an edge if and only if they co-occur in either (i) the same sentence, or (ii) two consecutive sentences, \textit{given the sentence pair is logically coherent}. The second condition is motivated by \citet{reinhart1980conditions}, who advocates that discourse coherence is affected by the connectivity of entities in consecutive sentences. Within the context window of two consecutive sentences, degree of continuity of the second sentence to the first one is a measure of linear, sequential coherence \citep{van1980semantics}. Low degree of sequential coherence signals discontinuity in discourse, warning that the key-entities occurring in consecutive sentence may not be semantically well-connected. 

Following \citet{duari2021ffcd}, we use BERT Next Sentence Prediction (NSP) score \citep{devlin2019bert} as a marker for sequential coherence. We compute NSP scores for all pairs of consecutive sentences in the section text and identify sentence pairs that have unusually low NSP scores using IQR-based statistical outlier detection method \citep{tukey1977exploratory}. We compute Five-number summary, which is a set of descriptive statistics consisting of the five important sample percentiles: the sample minimum ($min$), the lower quartile ($Q1$), the sample median ($Q2$), the higher quartile ($Q3$), and the sample maximum ($max$). Inter-quartile range (IQR) is defined as $Q3-Q1$. Following standard practice, we define the threshold for outliers at the lower end as $\lambda = Q1 - 1.5 * IQR$. Sentence pairs with BERT NSP scores greater than $\lambda$ are considered logically and sequentially coherent and the constituent key-entities in the two sentences are linked with edges. We ignore sentence pairs with NSP score below $\lambda$ for defining edge relations between key-entities as such sentence pairs connote lack of logical and sequential coherence. Each edge in the network is assigned co-occurrence frequency of the adjacent key-entities (nodes) in the section as weight. Proceeding this way, we construct undirected, weighted network $L$ corresponding to text in section $\mathbb S$. 

\begin{figure}[!htbp]
    \centering
    \begin{subfigure}{\columnwidth}
    \centering
        \includegraphics[scale = .3]{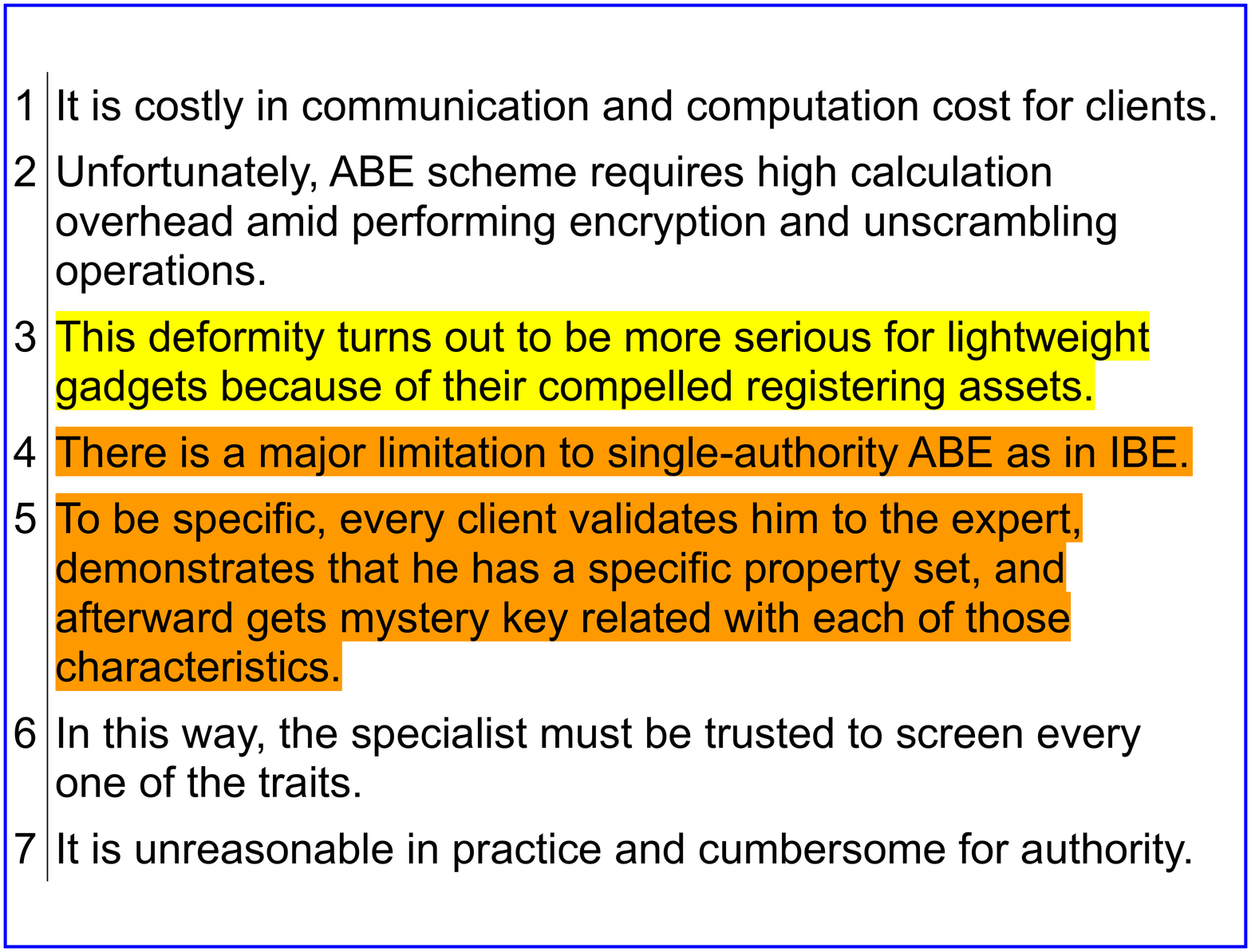}
        \caption{Sample text taken from document id `doc16' (section 4) from \textit{Neg} category.}
        \label{subfig:bad-text}
    \end{subfigure}
    \begin{subfigure}{\columnwidth}
    \centering
        \includegraphics[scale = .4]{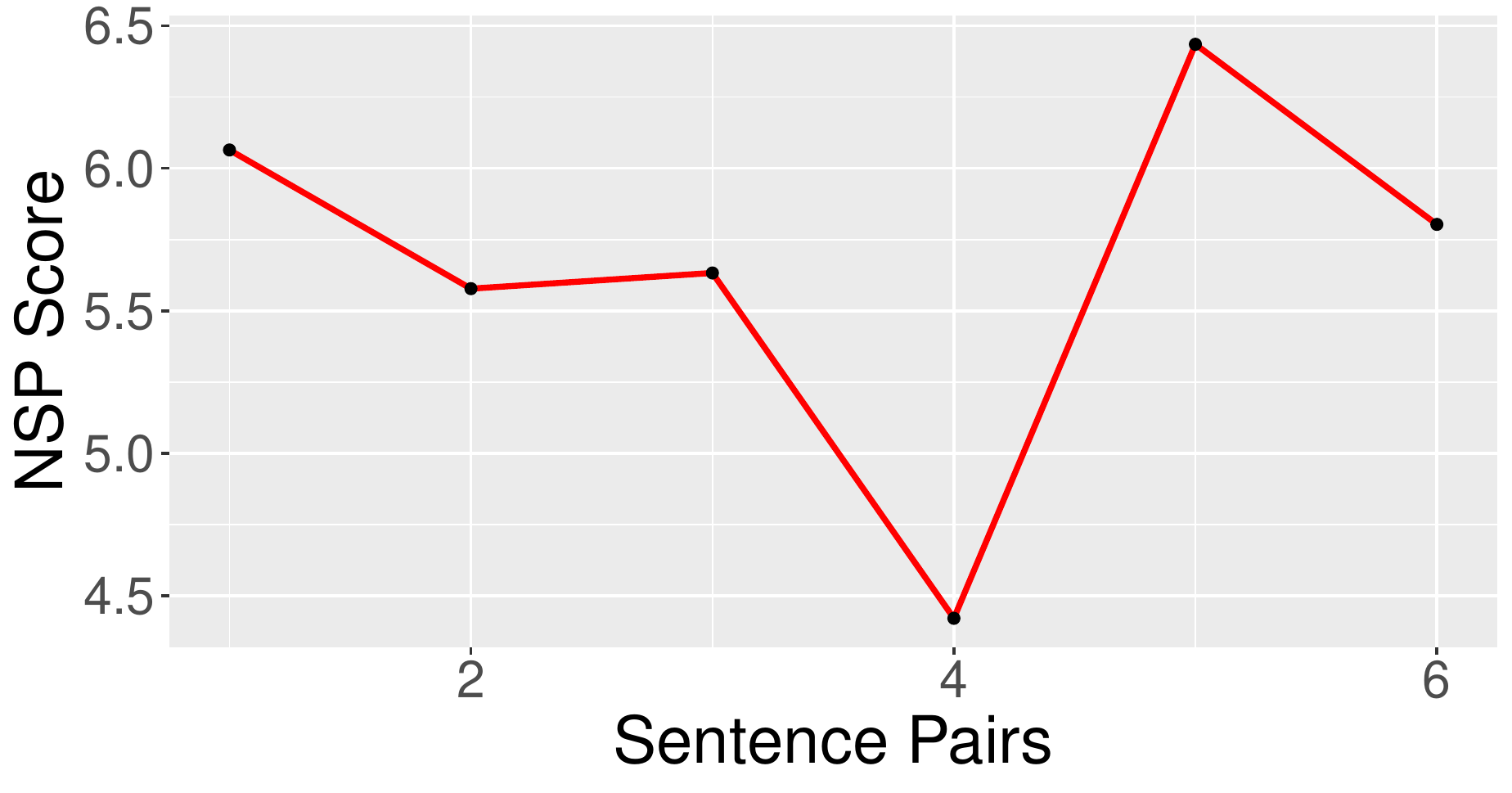}
        \caption{NSP scores for consecutive sentence pairs in Figure. \ref{subfig:bad-text}.}
        \label{subfig:coh-plot}
    \end{subfigure}
    \begin{subfigure}{\columnwidth}
    \centering
        \includegraphics[scale = .7]{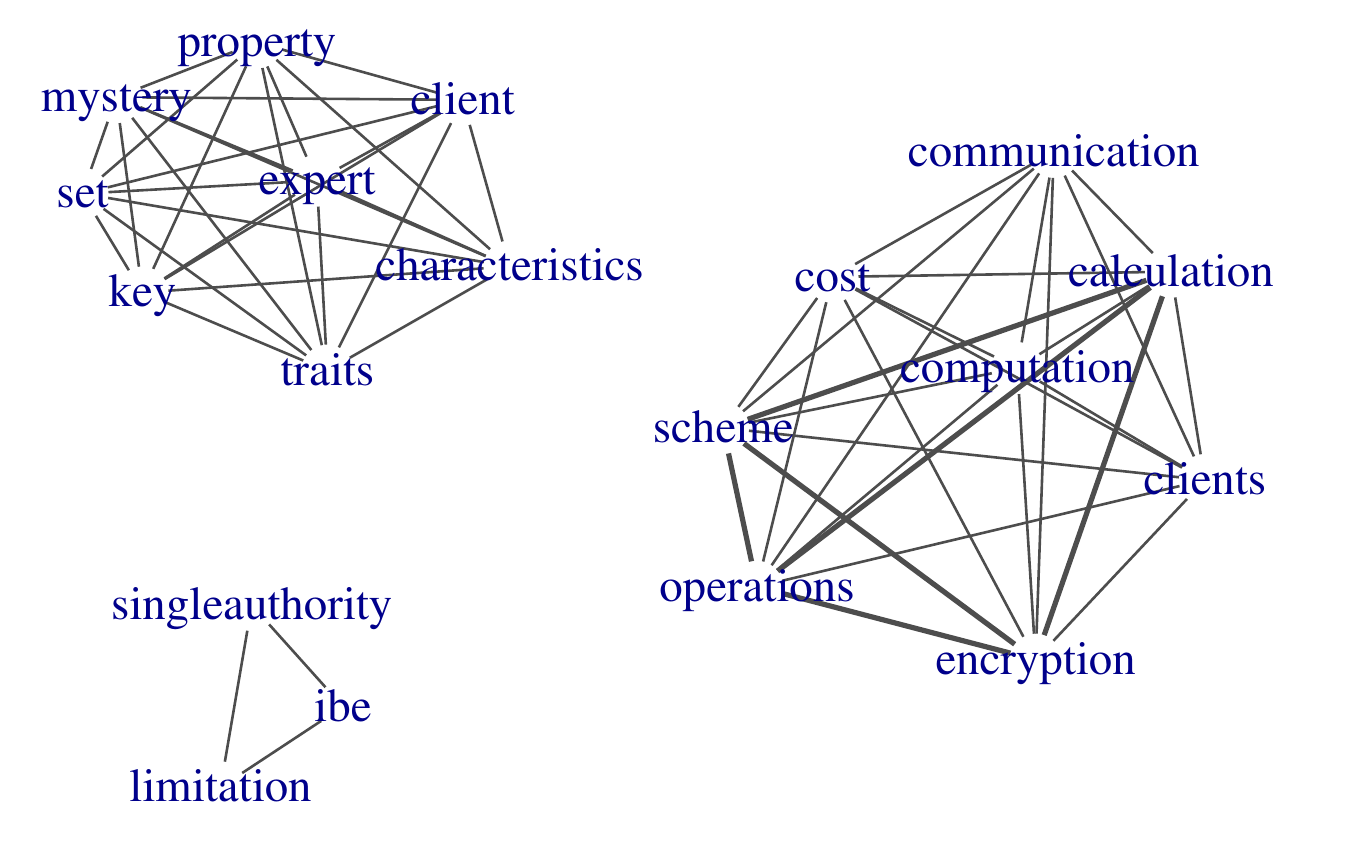}
        \caption{Graph for text in Figure. \ref{subfig:bad-text}.}
        \label{subfig:two-comp}
    \end{subfigure}
    \caption{Lack of coherence in writing resulting into multiple components in a network corresponding to a section.}
    \label{Fig:writing-quality}
\end{figure}

We present an illustrative example for construction of the network for a text in Figure \ref{Fig:writing-quality}. Given the sample text in Figure \ref{subfig:bad-text}, the BERT NSP score of each of the consecutive sentence pairs are plotted in Figure \ref{subfig:coh-plot}. 
The sentence pair $(4, 5)$ in Figure \ref{subfig:coh-plot} lacks sequential coherence due to which the entities present therein are not connected in the graph (Figure \ref{subfig:two-comp}). On the other hand, sentence id 3 - \textit{`This deformity ... registering assets.'}, despite being sequentially coherent with its immediate neighbors (NSP scores > 5.5 for pairs at positions 2 and 3 in Figure \ref{subfig:coh-plot}), does not contain any key-entities. Thus, key-entities from first two sentences are disconnected from the rest, which diminishes cohesion in the text. Such situations cause a impedance in discourse comprehension, which manifests as multiple components in the graph shown in Figure \ref{subfig:two-comp}. 

\subsection{Section-level Cohesion Metrics}
\label{subsec:sec-level-coh-metric}
As evident from the example in Figure \ref{Fig:writing-quality}, it is possible that a group of key-entities described in a section may be cohesive among themselves, but may not gel with other key-entities contained in the section. It can be caused by a text that contains two (or more) sets of sentences discussing either loosely related or unrelated sub-topics. Such situations are undesirable as they induce gaps in understanding, which may adversely affect the readers' comprehension. In this section, we discuss properties of the text which gauge the extent of cohesion.

We posit that smooth and coherent writing contains terms that co-occur in a similar context and results in network $L$ with a single component. Multiple components in the network $L$ indicate that the key entities in the discourse (section) are loosely linked, which signals lack of cohesion in the section text. This situation arises if a set of sequentially coherent sentences does not share key-entities with another set of sequentially coherent sentences in the section. For example, two sets of sequentially coherent sentences $\{1,2,3\}$ and $\{6,7\}$) in Figure \ref{subfig:bad-text} exhibit lack of cohesion among the entities. Key-entities in these two sets fall in disconnected components in the graph representation of the text (Figure \ref{subfig:two-comp}), and the author may like to review it to improve the cohesion.

The observation leads to the conjecture that the number of components in the network representation of text thus constructed is indicative of the quality of writing. Based on the premise that well-written texts exhibit lexical cohesion and thus result in a network with single component, while poorly-written texts exhibit cohesion gaps, which result in multiple components in the network, we hypothesize that the number of components in the network and the quality of writing are not independent. We test this hypothesis in Section \ref{subsec:res-multcomp} and present results.

\subsubsection{Section Level Cohesion Index}
It is accepted among linguists that stringing together and overlapping of phrasal units improve lexical cohesion \citep{stubbs2001computer, graesser2004coh, crossley2016tool}. Accordingly, we devise a metric, \textit{Section Level Index for Cohesion} (SLIC), to quantify cohesion in scientific scholarly texts by analyzing the corresponding network $L = \{V, E, W\}$. SLIC is based on average edge weight $w_{avg}$ in the network $L$, computed as
\setlength{\belowdisplayskip}{5pt} \setlength{\belowdisplayshortskip}{5pt}
\setlength{\abovedisplayskip}{5pt} \setlength{\abovedisplayshortskip}{5pt}
\begin{equation*}
\label{eq:}
  w_{avg} = \frac{1}{|E|} \sum_i \sum_j w_{ij}, i,j \in V.
\end{equation*}

We posit that a cohesive text is likely to exhibit high average edge weight, signifying a relatively stronger relationship between the nodes (key-entities). Furthermore, we expect a cohesive text to exhibit a relatively higher variance for the edge weights. This speculation is based on the premise that a cohesive text focuses on a central theme, with supplementary themes being referred to for completion. Key-entities related to the central theme are thus likely to share higher edge weights in contrast to entities that are related to the supplementary themes. We define SLIC as the coefficient of variation of edge weights as follows:
\setlength{\belowdisplayskip}{5pt} \setlength{\belowdisplayshortskip}{5pt}
\setlength{\abovedisplayskip}{5pt} \setlength{\abovedisplayshortskip}{5pt}
\begin{equation}
\label{eq:slic}
  SLIC = \frac{\sigma_W}{\mu_W}
\end{equation}

Here, $\mu_W$ and $\sigma_W$ are the mean and standard deviation of the edge weights in network $L$, respectively. SLIC score closer to $0$ indicates a higher probability of cohesion gaps in the text due to weak writing. SLIC scores can be greater than 1 when $\sigma_W > \mu_W$, which indicates high variation among the edge weights in $W$ signifying a discourse where key-entities pertaining to the central theme are strongly cohesive and key-entities from supplementary themes are moderately, yet sufficiently, connected to other key-entities in the network. We present empirical validation of $SLIC$ metric in section \ref{sec:experiments}.

\section{Cohesion at Document Level}
\label{sec:document-level}
After analyzing cohesion within sections of the text, we now proceed to scrutinize cohesion at the document level. This entails examining the quality of semantic interconnections between entities across sections. In the interest of simplicity, we consider transmission of ideas from one section to the next, connoting that the key-entities in the two successive sections are semantically related. Visualizing the complex network corresponding to each section as a layer, we establish semantic interconnections between consecutive sections and model the scientific scholarly text as a multilayer network. We first formalize the notation for the multilayer network below, and then describe the mechanism for creating interconnecting edges between layers. Subsequently, we describe the scheme to condense the network and glean quantitative measures that reveal cohesion at document level.
\subsection{Notation for Multilayer Network}
Following \citet{boccaletti2014structure}, we define multilayer network $\mathcal M = \{\mathcal L, \mathcal C\}$ as a tuple, with $\mathcal L$ denoting a family of graphs $ \mathcal L = \{L_1, L_2, \ldots, L_N\}$, called the layers, and edge-set $\mathcal C$ denoting interconnections between the layers. Layer $L_i = \{ V_i, E_i, W_i\}$ is a graph with $V_i$ denoting the set of nodes, and $E_i (\subseteq V_i \times V_i)$ denoting the set of edges. $W_i$ is the weighted adjacency matrix denoting weights of edges in $E_i$. Edges in $E_i$ are called intralayer edges. The layers are linked by a set of interlayer edges denoted by $\mathcal C = \{\mathbb {E}_{\alpha,\beta} \subseteq V_\alpha \times V_\beta; \alpha, \beta \in \{1, \ldots, N\}, \alpha \ne \beta \}$, each encoding the strength of relationship between nodes in two layers.

In the present context, scholarly document $D$ with $N$ sections is fragmented into $N$ discourse units (sections) and each section $\mathbb S_i$ is represented by layer $L_i$. Recall that the network for a section is an undirected, weighted graph (Section \ref{subsec:text-as-cn}). Nodes in $L_i$ are key-entities in the section and intralayer edges denote co-occurrence relation between the nodes. Layer $L_i$ is represented as an adjacency matrix $W_i$ of size $|V_i| \times |V_i|$, where element $w^{rs}_i$ denotes weight of the edge linking nodes $v^{r}_i \mbox{ and }v^{s}_i$ in layer $L_i$. Unconnected nodes have corresponding elements in $W_i$ set to zero. 

Interlayer edges in $\mathcal{M}$ portray semantic links between successive sections in the text. Interlayer edges or interconnections between layers are represented by $(N-1)$ weight matrices $C_{\alpha,\alpha+1}; \alpha \in \{1, \ldots, N-1\}$. Since we regard the flow of information across sections as sequential, element $c^{rs}_{\alpha,\alpha+1}$ is the weight of interlayer edge linking nodes $v^{r}_\alpha$ and $v^{s}_{\alpha+1}$ in layers $L_\alpha$ and $L_{\alpha+1}$, respectively. Note that interlayer edges are weighted and implicitly directed, simultaneously capturing the strength of semantic connections and flow of ideas between sections.
\subsection{Interconnecting Layers}
\label{subsec:inter-layer-network}
To capture the semantic relationship between key-entities in different layers, we derive their section specific contextual word embedding using BERT \citep{devlin2019bert}. A key-entity may occur multiple times and in different contexts within the section text. Ergo, we average the embeddings from all contexts in the section for the key-entity and represent it as a single node in the layer network. We compute pairwise cosine similarity between embeddings of all key-entities in two consecutive layers and use it as weight for the corresponding interlayer edges. Thus, weight of the interconnection linking nodes $v_\alpha^x$ and $v_{\alpha+1}^y$, denoted by $c_{\alpha, \alpha+1}^{xy}$, refers to the semantic similarity between the entities $v^x$ and $v^y$ in layers $L_\alpha$ and $L_{\alpha+1}$, respectively. Noting that the scores lie in the range $[-1, 1]$ and negative values are undesirable, we retain edges with similarity scores greater than or equal to 0.5. This threshold in the authors' view is a reasonable indicator of semantic strength required for understanding flow of thematic progression and the extent of lexical cohesion across layers (sections) of the scientific scholarly text.

\subsection{Condensing the Multilayer Network}
Scientific scholarly texts are typically long documents and generate numerous key-entities at each layer. However, a reader takes cognizance of the concepts presented in the text and their semantic relationships to appreciate the content. Concepts described in the text are thus important semantic units and their inter-connectivity across sections is instrumental for smooth thematic progression. Since concepts are often latent and not directly accessible, we identify groups of co-occurring words in the network at each layer and designate them as concepts as established in earlier research \citep{jia2018concept, paranyushkin2019infranodus}. Semantically binding together key-entities as a concept additionally serves the purpose of condensing the network at each layer, and down-sizing the interlayer edges. 
 
\begin{figure}[!htbp]
    \centering
    \begin{subfigure}{\textwidth}
    \centering
        \includegraphics[scale=.4]{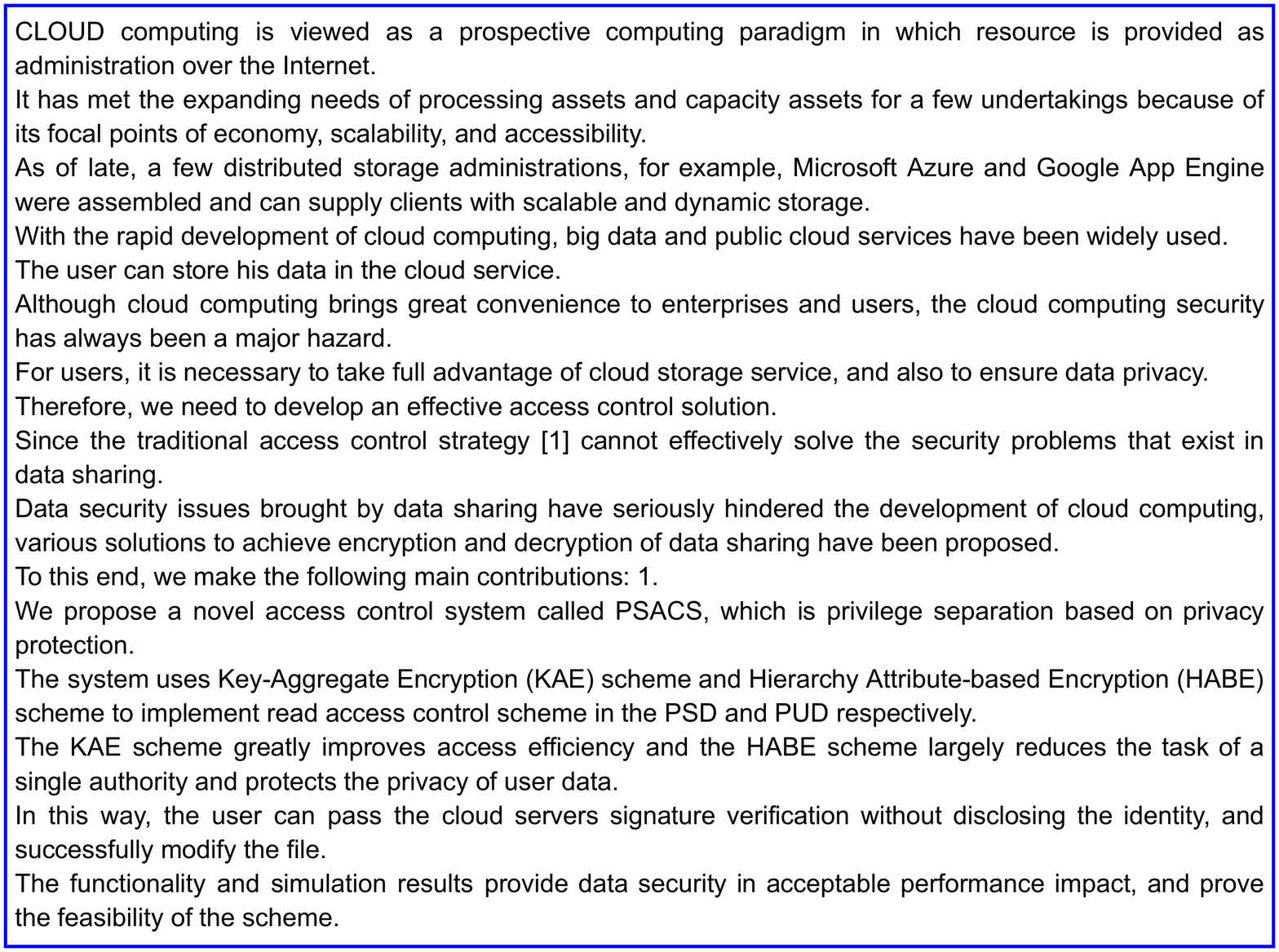}
        \caption{Sample section text.}
        \label{subfig:sample-text}
    \end{subfigure}
    \begin{subfigure}{0.6\textwidth}
    \centering
        \includegraphics[scale = .5]{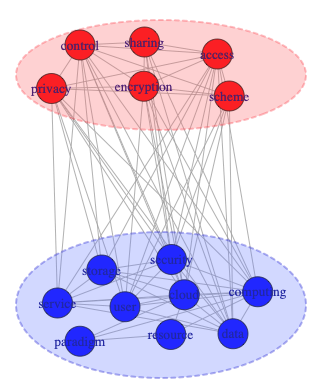}
        \caption{Layer network $L_i$ constructed from sample text in Fig. \ref{subfig:sample-text}. Nodes are key-entities extracted from the text. Two communities identified in the text encapsulate two concepts discussed in the text. }
        \label{subfig:comm-detect}
    \end{subfigure}%
    \hspace{0.3cm}
    \begin{subfigure}{0.3\textwidth}
    \centering
        \includegraphics[scale = .23]{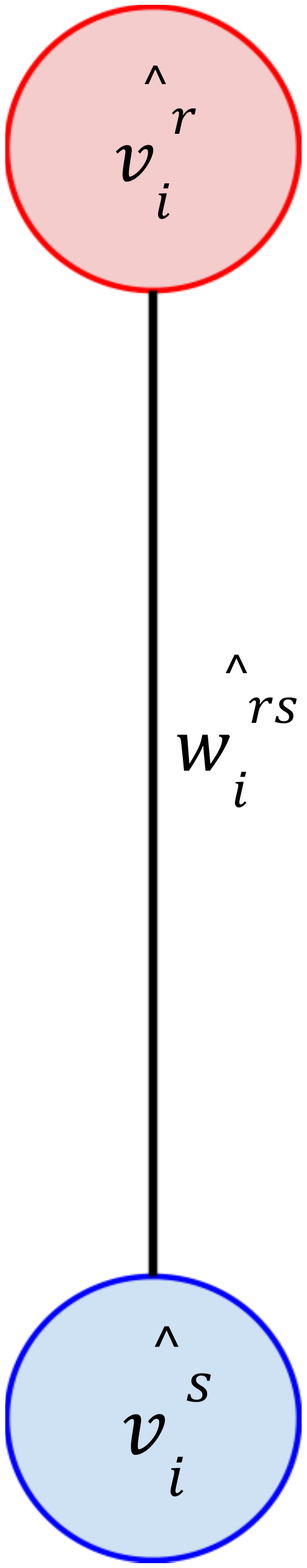}
        \caption{Layer metagraph $\hat L_i$ condensed from Fig. \ref{subfig:comm-detect}. Edge weight is computed as per Eq. \ref{eq:intralayer-metaedge-weight}.}
        \label{subfig:layer-metagraph}
    \end{subfigure}
    \caption{Modeling a section text as a layer network.}
    \label{fig:community-detection-example}
\end{figure}

In a network of text, communities portray a collection of semantically related words and proxy for concepts described in the section. We apply a community detection algorithm in each layer to discover concepts described in the corresponding section. We choose Louvain algorithm \citep{blondel2008fast} to detect communities because it does not require the number of communities as input, and is efficient. An example text with network abstracted as communities is presented in Fig. \ref{subfig:sample-text} and \ref{subfig:comm-detect}.

Notationally, the layer network $L_i$ corresponding to section $\mathbb S_i$ is transformed to metagraph $\hat L_i = (\hat V_i, \hat E_i)$, where community $r$ in $L_i$ is modeled as a meta-node ($\hat v_i^r \in \hat V_i$) in the transformed layer $\hat L_i$. Link between communities $r$ and $s$ (i.e., nodes $\hat v_i^r$ and $\hat v_i^s$) is denoted by the metaedge $ \hat e_i^{rs}$ in $\hat L_i$. We construct the appropriate weight matrix $\hat W_i$ for $\hat L_i$ based on the intuition that the weight of the metaedge $\hat e_i^{rs}$ connecting metanodes (communities) $\hat v_i^r$ and $\hat v_i^s$ is an additive function of weights of the intralayer edges between the constituent nodes, weighted by the number of connections between two. Accordingly, we define $\hat w^{rs}_i$ as shown below.
\setlength{\belowdisplayskip}{5pt} \setlength{\belowdisplayshortskip}{5pt}
\setlength{\abovedisplayskip}{5pt} \setlength{\abovedisplayshortskip}{5pt}
\begin{equation}
\label{eq:intralayer-metaedge-weight}
    \hat w^{rs}_i = log(\sum \sum w^{xy}_{i}) * m^{rs}
\end{equation}
where $x$ and $y$ are the key-entities belonging to communities $r$ and $s$ respectively, and $m^{rs}$ is the number of edges between the two communities. Two concepts that do not share co-occurring words have metaedge weight set to zero. Since the edge weights, and therefore the sum of edge weights, can be arbitrarily large, we apply a log-based transformation to scale down the large values.

To maintain sanity, compatible transformation of interconnections is obligatory. Recall that interlayer edges have weights that capture the semantic similarity between the corresponding entities. Accordingly, interlayer metaedge in the transformed space represent linkages between the concepts in two successive sections. Edge weight between two metanodes $\hat v^r_\alpha$ and $\hat v^s_{\alpha+1}$, in layers $\alpha$ and $\alpha+1$ respectively, is the sum of the weights of interlayer edges between the entities in communities $r$ and $s$. Formally, weight of the interlayer metaedge $ \hat e_{\alpha,{\alpha+1}}^{rs}$ is given by 
\setlength{\belowdisplayskip}{5pt} \setlength{\belowdisplayshortskip}{5pt}
\setlength{\abovedisplayskip}{5pt} \setlength{\abovedisplayshortskip}{5pt}
\begin{equation}
    \hat c_{\alpha,{\alpha+1}}^{rs} = \sum \sum{c_{\alpha, \alpha+1}^{xy}} \, ,
\end{equation}
where $c_{\alpha, \alpha+1}^{xy}$ is the weight of the interlayer edge linking $v_\alpha^x (\in \hat v_\alpha^r)$ and $v_{\alpha+1}^y (\in \hat v_{\alpha+1}^s)$. It is noteworthy that all metanodes in layer $\hat L_\alpha$ are connected to all metanodes in $\hat L_{\alpha + 1}$. Thus, excruciating details of multilayer representation $\mathcal M$ of document $D$ are condensed as compact multilayer network $\mathcal{\widehat M}$, which can be interpreted with ease. 

\begin{sidewaysfigure}[!htbp]
    \centering
        \includegraphics[scale = 0.6]{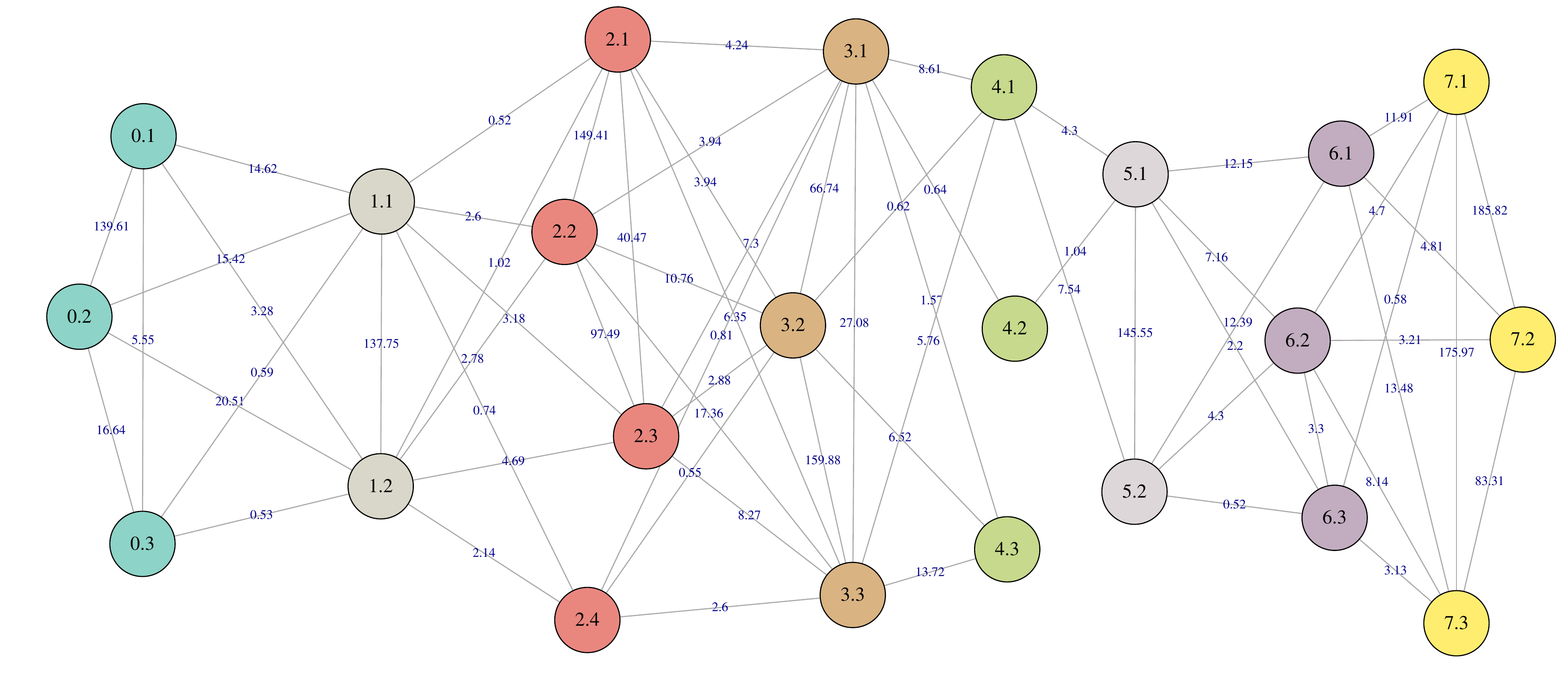}
        \caption{Modeling a document as multilayer network. We present metagraph $\mathcal{\widehat M}$ of document id `doc16' from `Neg' category, with concepts as metanodes connected by intralayer and interlayer metaedges. Nodes in same color are the concepts in same section (layer), with vertex label in the format (\texttt{i.r}) denoting  community (concept) $\hat v_i^r$ in layer network $\hat L_i$. Intralayer metaedges denote consolidated co-occurrence weight between the concepts and interlayer metaedges denote aggregated semantic similarity between the concepts discussed in consecutive sections of the document.}
        \label{fig:community-metagraph}
\end{sidewaysfigure}

Figure \ref{fig:community-detection-example} demonstrates an example transformation from layer $L_i$ to $\hat L_i$. The text excerpt shown in Figure \ref{subfig:sample-text} is taken from a random section of document id `doc16' from `Neg' category in the experimental dataset. The section text is transformed into a complex network (layer $L_i$), where the nodes are key-entities present in the text and the links denote co-occurrence relationship among the key-entities. We apply community detection algorithm on $L_i$ and detect two communities as shown in Figure \ref{subfig:comm-detect}, which are encircled in red and blue color. The two communities are then transformed into two metanodes ($\hat v_i^r$ and $\hat v_i^s$), which are connected with an metaedge ($\hat e_i^{rs}$) with weight $\hat w^{rs}_i$ computed as equation \ref{eq:intralayer-metaedge-weight} (Figure \ref{subfig:layer-metagraph}). We process each section of the arbitrary document in a similar way and create the MLN metagraph $\mathcal{\widehat M}$. Figure \ref{fig:community-metagraph} presents the condensed multilayer network $\mathcal{\widehat M}$, where interlayer metaedges denote semantic similarity between pairs of concepts (communities) discussed in two consecutive sections.


\subsection{Document Level Cohesion Metrics}
\label{subsec:cohesion}
MLN representation of a target document $D$ encodes features of the scientific scholarly writing, which effectively convey cohesion of the semantic units and consistency of ideas developed in the text. We identify select network properties from the multilayer network $\mathcal{M}$ and its condensed version $\mathcal{\widehat M}$, and ascribe them to cohesion at global level. We derive global cohesion metrics from network properties of the multilayer network representation that aid in identifying cohesion gaps in the document.

We follow \citeauthor{cancho2001small}'s (\citeyear{cancho2001small}) study, where the authors found that word co-occurrence networks in English language exhibit small-world property \citep{watts1998collective} with small average shortest path length and a high clustering coefficient. We build on this assertion and posit that each layer network $L_i$ in $\mathcal{M}$, which are based on co-occurrence relations between key-entities, are likely to exhibit strong small-world characteristics proportionate to the degree of cohesion in the corresponding text. Accordingly, we define four metrics - the first two inspired from small world \citep{watts1998collective} property are based on the layer networks in $\mathcal{M}$, and the other two are based on the condensed network $\mathcal{\widehat M}$.

\subsubsection{Entity Connectivity Index}
We expect that a section text with high degree of lexical cohesion would result into a dense network representation. Therefore, a well-written scientific scholarly text with high degree of lexical cohesion is expected to exhibit a dense network at all layers corresponding to the sections in the document. Consequently, the layers in $\mathcal{M}$ should individually exhibit high weighted average clustering coefficient (WCC). We adopt \citet{barrat2004architecture} definition of weighted average clustering coefficient,and present the computation below. Given a network $G = \{V, E, W\}$, the weighted clustering coefficient $wc_j$ for node $v_j \in V$ is computed as:
\setlength{\belowdisplayskip}{5pt} \setlength{\belowdisplayshortskip}{5pt}
\setlength{\abovedisplayskip}{5pt} \setlength{\abovedisplayshortskip}{5pt}
\begin{equation*}
wc_j = \frac{1}{s_j(k_j-1)} \sum_{h,k} \frac{(w_{jh}+w_{jk})}{2} a_{jh} a_{hk} a_{jk},
\end{equation*}

where $k_j$ and $s_j$ are the degree and strength (weighted degree) of node $v_j$, respectively. The indicator variable $a_{jh}$ is $1$ if there exists an edge between nodes ($v_j, v_h$), $w_{jh}$ being the weight of the edge. The computation for a triad $v_j$, $v_h$, and $v_k$ is non-zero only when there exist a triangle centered at node $v_j$. The weighted average clustering coefficient (WCC) for network $G$ with $n$ nodes is defined as:
\setlength{\belowdisplayskip}{5pt} \setlength{\belowdisplayshortskip}{5pt}
\setlength{\abovedisplayskip}{5pt} \setlength{\abovedisplayshortskip}{5pt}
\begin{equation*}
WCC = \frac{\sum_j wc_j}{n}.
\end{equation*}

Using this computation, we calculate $WCC_i$ for layer networks $L_i$ (Section $\mathbb S_i$) in the network. We define Entity Connectivity Index (ECI) of the text, which quantifies text cohesion in terms of global clustering coefficient as follows:
\setlength{\belowdisplayskip}{5pt} \setlength{\belowdisplayshortskip}{5pt}
\setlength{\abovedisplayskip}{5pt} \setlength{\abovedisplayshortskip}{5pt}
\begin{equation}
  {ECI} = \sqrt{\sum_i \frac{(1-WCC_i)^2}{N}},
\end{equation}

\noindent
where $N$ is the number of layers in $\mathcal{M}$ (sections in the text). $ECI$ quantifies the root mean squared deviation of weighted clustering coefficients from the ideal value of $1$. Value of the metric lies in the range $[0,1]$ and a well-written, cohesive text exhibits $ECI \approx 0$. Extreme $ECI$ value of 1 indicates a situation where no triad forms an triangle, and no three entities are co-occurring in same or adjacent sentences. Such written discourse is clearly poor, and requires refinements in writing to address the cohesion gaps.

\subsubsection{Entity Proximity Index}
We further posit that a cohesive section text with well-knit discourse entities would result into a network ($L_i$) with low average shortest path length, which is a characteristic of small-world networks. Since average shortest path length (APL) measures the average number of edges separating any two nodes in the network, it is a marker of semantic proximity between the entities in the current context. The underlying premise is that cohesive scholarly writings describe contextually similar concepts with mild and natural variations in semantic relatedness. Therefore, the constituting entities frequently co-occur and are expected to maintain sufficient semantic relatedness to form a coherent narrative. According to small-world property described in the seminal paper by \citet{watts1998collective}, the average shortest path length of the network is expected to be significantly lower than the number of nodes in the network $\left ( APL_i \propto \log(n_i) \right )$. We compute APL for network $G$ as follows \citep{zaki2014data}:
\setlength{\belowdisplayskip}{5pt} \setlength{\belowdisplayshortskip}{5pt}
\setlength{\abovedisplayskip}{5pt} \setlength{\abovedisplayshortskip}{5pt}
\begin{equation*}
APL = \frac{2}{n (n - 1)} \sum_j \sum_{k>j} d(v_j, v_k),
\end{equation*}
where $n$ is the number of nodes in $G$ and $d(v_j, v_k)$ is the shortest distance (i.e. proximity) between nodes $v_j$ and $v_k$. In case of disconnected networks where $d(v_j, v_k)$ is undefined, we set $d(v_j, v_k)$ to $n$. We define Entity Proximity Index (${EPI}$) to connote relatedness of the entities described in the section as follows:
\setlength{\belowdisplayskip}{5pt} \setlength{\belowdisplayshortskip}{5pt}
\setlength{\abovedisplayskip}{5pt} \setlength{\abovedisplayshortskip}{5pt}
\begin{equation}
  {EPI} = \sqrt{\sum_i \frac{(\log(n_i)-APL_i)^2}{N}},
\end{equation}
where $APL_i$ is the average shortest path length of the i-th layer network $L_i$ in $\mathcal{M}$ and $n_i$ is the number of nodes in $L_i$. $EPI$ computes the average root mean squared deviation from the ideal average path length in a small-world network. The value of $EPI$ can be arbitrarily large depending on the quality of writing. However, in a well written and cohesive text where the entities are repeated deftly, the $EPI$ values are expected to be small and ideally close to 0.

\subsubsection{Concept Connectivity Index} 
We explore subgraph structures in the condensed network $\mathcal{\widehat M}$ to gauge the extent of cohesion between concepts described within and across sections of a scientific scholarly text. A well-written cohesive text is expected to exhibit topologically near-complete subgraphs in $\mathcal{\widehat M}$ that connect the concepts (metanodes) in the network. 

\begin{figure}[!htbp]
    \centering
    \includegraphics[scale=3]{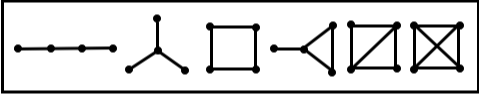}
    \caption{All possible subgraphs of size 4 - considering each node as a concept and edge as semantic relation between concepts, the fully connected subgraph ($K4$) signifies cohesive writing.}
    \label{fig:size-4-graphlets}
\end{figure}

We follow \citeauthor{bondy1976graph}'s (\citeyear{bondy1976graph}) definition of a \textit{subgraph}, which states that a graph $H = \{V', E', W'\}$ is a subgraph of $G = \{V, E, W\}$ (written as $H \subseteq G$) if $V' \subseteq V$, $E' \subseteq E, E' \subset V' X V'$ and $W' \subseteq W$. A complete subgraph, where all constituent nodes are connected to all other nodes, is a region of high density, which translates to strongly cohesive writing in the context of discourse analysis. Accordingly, the count of complete subgraphs in the network representation of text beacons the extent of cohesion in writing. We count complete subgraphs of size four (K4) to indicate cohesion among concepts (Fig. \ref{fig:size-4-graphlets}). We chose subgraphs of size 4 in our analysis as we believe smaller sizes (three or less) are deficient in capturing cohesion, while bigger sizes (five or more) are computationally expensive to find. 

It is noteworthy that by construction, the concepts (metanodes) in the network $\mathcal{\widehat M}$ are fully-connected - both within and across layers. We prune metaedges from $\mathcal{\widehat M}$ to discard weak connections between concepts, both within and across layers. Pruning is based on the premise that in cohesive writing, intralayer and interlayer metaedges in $\mathcal{\widehat M}$ have reasonably high weights signifying consistently strong semantic connections between the metanodes (concepts) within and across sections. However, texts with weak cohesion may contain metaedges with low weights that symbolize weak semantic connections. Accordingly, we define data-driven thresholds (as in Section \ref{subsec:text-as-cn}) separately for intralayer and interlayer metaedges to eliminate weak semantic links that manifest as outliers in metaedge weights distribution. The outliers help the author to localize concepts that are weakly connected within and across sections. Removal of metaedges (from $\mathcal{\widehat M}$) that have weight below the corresponding threshold uncovers weak semantic connections between the concepts in the document.

Removal of an edge between concepts, a pathology for weak cohesion, is reflected as reduced number of $K4$ in $\mathcal {\widehat M}$ after metaedge pruning. Therefore, decrease in the count of $K4$ after removing weak links is a candid statement of weak text cohesion. Let $K4_{bef}$ denote the number of $K4$ in $\mathcal {\widehat M}$, and $K4_{aft}$ denote the same after removal of weak meta-edges. We compute Concept Connectivity Index, $CCI$, as given below:
\setlength{\belowdisplayskip}{5pt} \setlength{\belowdisplayshortskip}{5pt}
\setlength{\abovedisplayskip}{5pt} \setlength{\abovedisplayshortskip}{5pt}
\begin{equation}
    CCI = 1 - \frac{K4_{aft}}{K4_{bef}} 
\end{equation}
The metric value lies in the range $[0,1]$. If no meta-edge is removed from $\mathcal {\widehat M}$, $CCI$ takes the ideal value of $0$ implying that the text consistently exhibits smooth and strong linkages between the concepts in the text, both within and across sections. As the $CCI$ value gets farther from 0, the semantic connectivity among the concepts in the scientific scholarly text gets weaker.

\subsubsection{Isolated Concepts Index} 
Existence of isolated concepts in section, i.e., metanodes with no link to other metanodes in $\hat L_i$ after metaedge pruning suggests a lack of cohesion in the corresponding section-level text. Thus, existence of isolated metanodes in the layer networks $\hat L_i$ of $\mathcal{\widehat M}$ is an additional indicator of lack of lexical cohesion. We define Isolated Concepts Index, ${ICI}$, as a metric for document-level cohesion by aggregating the section-level indicators as shown below (Eq. \ref{eq:ici}). Formally, if $\hat n_i$ denotes the number of metanodes in layer $\hat L_i$ and $\hat{m_i}$ denotes the number of isolated metanodes in layer network $\hat L_i$ in a network $\mathcal{\widehat M}$ with $N$ layers, then 
\setlength{\belowdisplayskip}{5pt} \setlength{\belowdisplayshortskip}{5pt}
\setlength{\abovedisplayskip}{5pt} \setlength{\abovedisplayshortskip}{5pt}
\begin{equation}\label{eq:ici}
  {ICI} = \frac{\sum_i \hat{m_i}}{\sum_i \hat n_i}
\end{equation}

The value of ${ICI}$ lies in the range $[0,1]$. Well written, cohesive text has ${ICI} = 0$, indicating good semantic connectivity between concepts described in the scientific scholarly text. This metric also facilitates localization of weak cohesion within and across sections. The author may either add or rephrase the content to bind entities in the related concepts for smooth comprehension of the text.

\section{Dataset and Experimental Design}
We design experiments to provide proof-of-concept for the proposed multilayer networks model for text representation and \textit{CHIAA} framework. The study focuses on assessing effectiveness of the proposal to capture thematic progression of discourse and lack of lexical cohesion in the text.

\subsection{Datasets}
\label{sec:data}
We use two data collections in our experiments. The first collection comprises 172 articles accepted in ICLR 2017 and is a part of the PeerRead collection \citep{kang18peerread}. The second collection is a curated dataset of 100 articles published in predatory venues. 

\vspace{0.5cm}
\noindent
\textbf{\textit{ICLR2017\_ACC} dataset\label{para:ch6-data1}:} The ICLR2017 dataset is a part of the PeerRead collection \citep{kang18peerread}, which contains 427 papers submitted to ICLR2017 conference with corresponding accept/reject decisions. Out of these 427 papers, we use the 172 accepted papers (\textit{ICLR2017\_ACC}) in our experiments as examples of good quality writing (Category \textit{Pos}). We refrain from using the remaining 255 rejected articles as the set of contrasting examples of poor-quality writing because reject decision is made based on multiple factors - such as lack of technical soundness/correctness or novelty, inadequate experiments, etc., and not solely based on writing quality. A manuscript can be well-written, yet it may lack technical soundness to impress all reviewers.

\vspace{0.5cm}
\noindent
\textbf{\textit{Predatory texts}\label{para:ch6-data3}:} We curate this dataset for complementing the selected articles from ICLR2017\_ACC collection by collecting 100 randomly selected articles published in predatory journals\footnote{Randomly-picked from \href{https://beallslist.net/standalone-journals/}\texttt{https://beallslist.net/standalone-journals/}.} (Category \textit{Neg}), published during 2011-2020\footnote{The dataset and code will be made available at the corresponding author's GitHub repository}. We use this dataset as a contrasting example of weak scholarly writing compared to the ICLR2017\_ACC articles in our empirical analysis.

\noindent
We perform basic cleaning of the text in both collections by removing equations, section and subsection headings, tabular data, captions, etc. We present dataset statistics after cleaning in Table \ref{tab:dataset-details}. We posit that articles in \textit{Pos} category are well-written and are thus more cohesive and consistent with the context. On the contrary, articles published in \textit{Neg} category \textit{may} have relatively subpar writing quality due to relaxed standard of writing. We analyze the documents in the two datasets and present proof-of-concept for section level and document level cohesion analysis.

\begin{table}[!htbp]
    \centering
    \begin{tabular}{c|c|c|c|c}
    \hline
        \textbf{Dataset} & $\mathbf{|D|}$ & $\mathbf{|Sec|}$ & $\mathbf{Sec/Doc}$ & $\mathbf{Sent/Sec}$\\ \hline
        ICLR2017\_ACC (Pos) & 172 & 1175 & 6.735 & 33.683\\
        Predatory texts (Neg) & 100 & 649 & 6.49 & 23.325 \\ \hline
    \end{tabular}
    \caption{Dataset statistics after cleaning. $\mathbf{|D|}$: Number of documents, $\mathbf{|Sec|}$: total number of sections in the category, Sec/Doc: average number of sections per document, and Sent/Sec: average number of sentences per section.}
    \label{tab:dataset-details}
\end{table}

\subsection{Research Questions}
\label{subsec:research-questions}
We design our experiments to answer the following research questions.
\begin{enumerate}
    \item \textit{Do multiple components in the network representation of text signify cohesion gaps?\\} 
    We extract section texts from each document in our dataset and label them appropriately as `Pos' and `Neg' based on the category of the parent document. We construct section networks (Section \ref{subsec:text-as-cn}) and analyze the number of components in the networks of each class. We use Chi-square test to check if ``Number of Components” and ``Document category” are independent, and present our analysis in section \ref{subsec:res-multcomp}.

    \item \textit{How well does $SLIC$ capture section level coherence?\\} 
    We analyze the sections individually from each document of the experimental dataset and accordingly label the constituent sections in each document as `Pos' and `Neg' categories. We compute $SLIC$ score for each section, and check the correlation of the score with established cohesion metrics derived from TAACO tool \citep{crossley2016tool,crossley2019tool}. We flag sections with low $SLIC$ scores and localize the regions of text that attribute to cohesion gap for rewriting. Results affirming our conjecture are presented in Section \ref{subsec:SLIC_vs_TAACO}.
    
    \item \textit{How well does the proposed document-level metrics capture (lack of) global cohesion in the text?\\} 
    We design an experiment to analyze the documents in the dataset and evaluate the effectiveness of the proposed metrics in capturing global lexical cohesion. We compute $ECI$, $EPI$, $CCI$ and $ICI$ metrics from each documents in the dataset, and check the correlation of these scores with our hypothesis that documents from `Pos' category exhibit lower scores (closer to $0$) for these measures as they are likely to be more cohesive than `Neg' category documents. We believe that individually cohesive sections may not always result in a cohesive text when combined. Thus, we investigate the extent of global cohesion using these four metrics and flag documents that exhibit high scores. Results affirming our assumption are presented in Section \ref{subsec:K4-APL-res}.
\end{enumerate}

\section{Experimental Results and Analysis}
\label{sec:experiments}
In this section, we present empirical analysis to answer the research questions posed in section \ref{subsec:research-questions}. We validate the proposed section-level and document-level cohesion metrics and establish their efficacy for distinguishing two categories of scientific scholarly texts.

\subsection{Multiple Components in a Network}
\label{subsec:res-multcomp}
We test our conjecture that multiple components in a section network is a distinctive feature of `Neg' category documents (Section \ref{sec:data}). The `Pos' category documents are generally considered well-written and `Neg` category documents are more likely to exhibit cohesion gaps. We extract the section texts from each document in the two datasets and construct a network for each section. The nodes (key-entities) are identified from the corresponding section text using a complex network based keyword extractor \citep{duari2020complex}. We apply Chi-square test for independence of the observed variable ``Components” against the variable ``Document category” which can have label `Neg' or `Pos' (Section \ref{subsec:sec-level-coh-metric}). We test the hypothesis 
\setcounter{hyp}{-1}
\begin{hyp} \label{hyp:null}
\textit{There is no association between the variable ``Document category” and the observed variable ``Components”.}
\end{hyp}

\noindent
against the alternative hypothesis,

\begin{hyp} \label{hyp:alt}
\textit{The variable ``Document category” is contingent upon the observed variable ``Components”.}
\end{hyp}

As evident from Table \ref{tab:dataset-details}, the number sections in `Pos' and `Neg' categories are imbalanced. Even though Chi-square test is agnostic to imbalance in categories, in the interest of robustness, we perform Chi-square test on three variations of data - 
\begin{enumerate}
    \item all sections from `Pos' category documents ($1175$) and `Neg' category documents ($649$),
    \item 100 `Neg' documents and 100 randomly sampled `Pos' documents with all their respective sections, and
    \item 649 sections from `Neg' documents and a equal number of randomly sampled sections from `Pos' documents. 
\end{enumerate} 

The contingency tables for the three variations are shown in Tables \ref{subtab:all_sec}, \ref{subtab:100_doc} and \ref{subtab:eq_sec}, respectively. 
\begin{table}[!htbp]
    \begin{subtable}[t]{0.48\columnwidth}
    \centering
    \begin{tabular}{c| c c}
        \hline
        \multirow{2}{*}{\textbf{Components}} & \multicolumn{2}{c}{\textbf{Category}}\\
        \cline{2-3}
         & Neg & Pos \\ \hline
        Multiple & 101 & 42 \\
        Single & 548 & 1133 \\ \hline
    \end{tabular}
    \caption{All sections from both categories ($p < 0.001$).}
    \label{subtab:all_sec}
    \end{subtable}%
    \hspace{\fill}
    \begin{subtable}[t]{0.48\columnwidth}
    \centering
    \begin{tabular}{c| c c}
        \hline
        \multirow{2}{*}{\textbf{Components}} & \multicolumn{2}{c}{\textbf{Category}}\\
        \cline{2-3}
         & Neg & Pos \\ \hline
        Multiple & 101 & 20 \\
        Single & 548 & 382 \\ \hline
    \end{tabular}
    \caption{100 documents from each category and the corresponding sections ($p < 0.001$).}
    \label{subtab:100_doc}
    \end{subtable}
    \begin{subtable}[t]{\textwidth}
    \centering
    \vspace{0.5cm}
    \begin{tabular}{c| c c}
        \hline
        \multirow{2}{*}{\textbf{Components}} & \multicolumn{2}{c}{\textbf{Category}}\\
        \cline{2-3}
         & Neg & Pos \\ \hline
        Multiple & 101 & 22 \\
        Single & 548 & 627 \\ \hline
    \end{tabular}
    \caption{Equal number of sections from both category ($p < 0.001$).}
    \label{subtab:eq_sec}
    \end{subtable}
    
    \caption{Contingency tables for Chi-square test for independence of variables.}
    \label{tab:contingency_tab}
\end{table}
We observe that the \textit{p}-values for the Chi-square tests on all three variations of data are $< 0.001$, implying that the null hypothesis is rejected in all three cases. It is, therefore, reasonable to conclude that section networks with multiple components are more likely to be found in negative category documents than positive ones.

We acknowledge that all sections from `Pos' category documents are not equally well written, and likewise, all sections from `Neg' category documents are not poorly written. Both category documents are likely to have sections with anomalous behavior. We, therefore, strengthen our claim with another observation that `Pos' category sections exhibit a relatively lower empirical probability of having section networks with multiple components. The observed probability for `Pos' category is $\frac{42}{1175} = 0.03574468$, i.e. $42$ sections have multiple components out of a total of $1175$ sections, while, `Neg' category sections exhibit a relatively higher probability of $\frac{101}{649} = 0.155624$, i.e. $101$ sections have multiple components out of $649$ sections.

\subsection{Correlation of SLIC Scores with Established Measures of Cohesion}
\label{subsec:SLIC_vs_TAACO}
We compare SLIC with established indices that measure cohesion at the text level. A variety of indices are reported in Coh-Metrix \citep{graesser2004coh,graesser2011coh} and TAACO \citep{crossley2016tool, crossley2019tool} tools for cohesion analysis. Since lexical cohesion concerns referencing of entities through repetition and synonymy-based relations, TTR- based and repeated content-based indices best capture this type of cohesion at the section level text. TTR-based indices are computed as the ratio of unique words (types) in a text to the total number of words (tokens) in the text. A high value of TTR ($\approx 1$) indicates that almost all tokens are unique, thus eliminating the repetition factor. Thus, we speculate that SLIC scores (Eq. \ref{eq:slic}) and TTR-based indices are negatively correlated. On the other hand, repeated content based indices exhibit a high value when there is a relatively higher degree of repetition, which indicates a higher amount of lexical cohesion. Therefore, we expect a positive correlation between SLIC and repeated content based indices. \citet{crossley2016tool} demonstrate positive relation of these selected indices with human-annotated measures of cohesion. We briefly describe the selected cohesion indices in Table \ref{tab:coh-taaco}.

\begin{table}[!htbp]
    \centering
    \begin{tabular}{p{6.5cm}|p{8cm}}
    \hline
        \textbf{Indices} & \textbf{Description} \\ \hline
        \texttt{lemma\_ttr} & Type-token ratio (TTR) computed after lemmatizing the tokens present in the text.\\ \hdashline
        \texttt{content\_ttr} & TTR using only content words.\\ \hdashline
        \texttt{repeated\_content\_lemmas} & The ratio of lemmatized content words being repeated in the sentences of the text. \\ \hdashline
        \texttt{repeated\_content\_and\_pronoun\_lemmas} & The ratio of lemmatized content words along with pronouns being repeated in the sentences of the text.\\ \hline
    \end{tabular}
    \caption{Details of relevant cohesion indices computed using TAACO 2.0 tool \protect{\citep{crossley2019tool}}.}
    \label{tab:coh-taaco}
\end{table}

We analyze sections from articles in `Pos' and `Neg' category documents with the restriction that the sections contain at least six sentences. We also discard the sections whose corresponding network contains less than four nodes (key-entities). We believe that such texts seldom merit a separate section. We observed that majority of these discarded sections belong to articles published in predatory venues. Finally, we are left with $1056/1175$ sections from `Pos' category and $503/649$ from `Neg' category. We compute SLIC scores for these sections and compare them against the above mentioned TAACO indices computed using recent TACCO 2.0 tool \citep{crossley2019tool}\footnote{Downloaded from hosting site at \href{https://www.linguisticanalysistools.org/taaco.html}\texttt{https://www.linguisticanalysistools.org/taaco.html}.}.

\begin{figure}[!htbp]
    \centering
    \begin{subfigure}{0.48\columnwidth}
    \centering
    \includegraphics[scale=0.4]{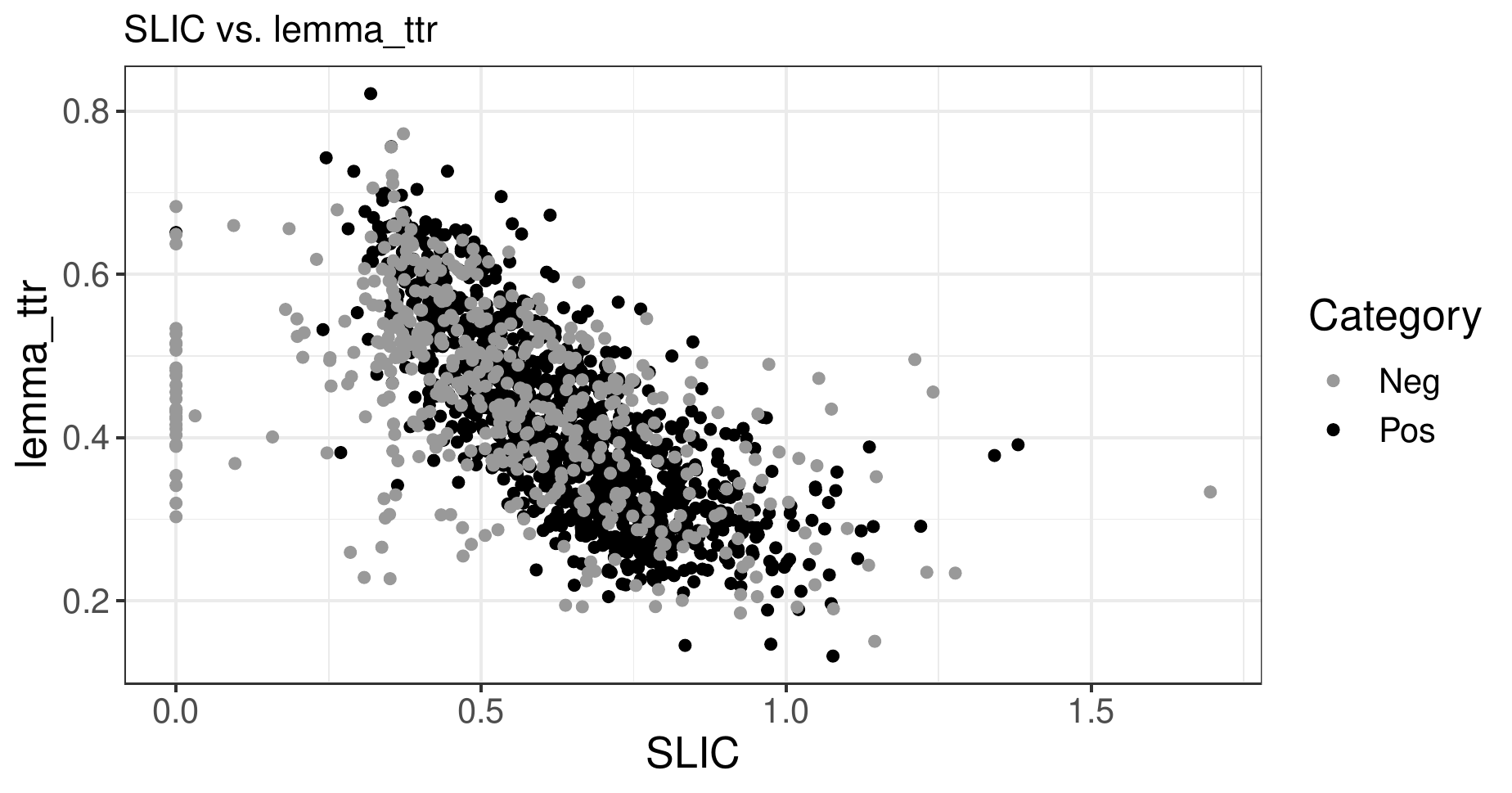}
    \caption{Correlation plot of SLIC vs. \texttt{lemma\_ttr}. Pearson Correlation Coefficient $\approx -0.637$.}
    \label{fig:slic-vs-lemma-ttr}
    \end{subfigure}%
    \hspace{\fill}
    \begin{subfigure}{0.48\columnwidth}
    \centering
    \includegraphics[scale=0.4]{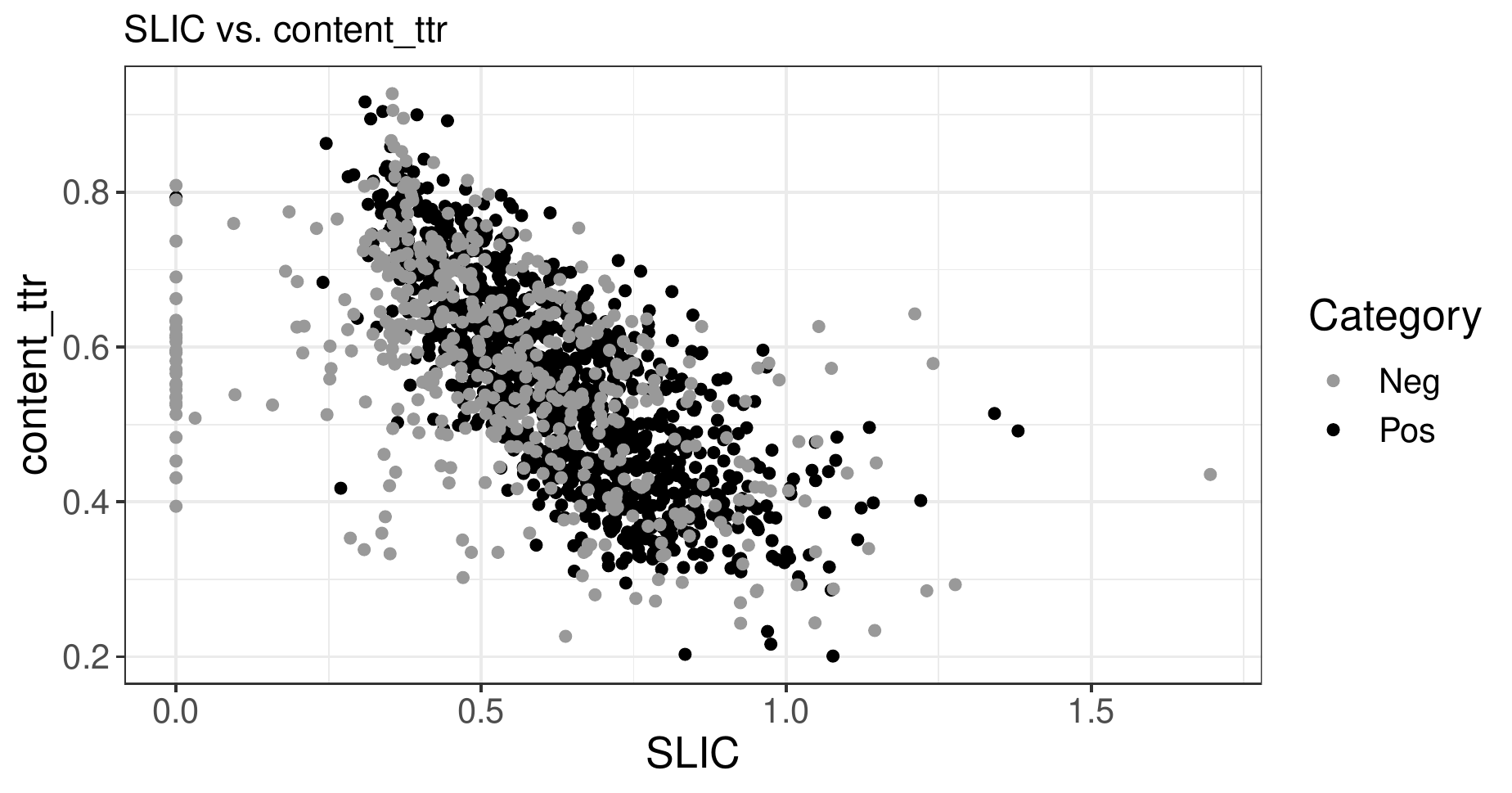}
    \caption{Correlation plot of SLIC vs. \texttt{content\_ttr}. Pearson Correlation Coefficient $\approx -0.63$.}
    \label{fig:slic-vs-content-ttr}
    \end{subfigure}
    \begin{subfigure}{0.48\columnwidth}
    \centering
    \includegraphics[scale=0.4]{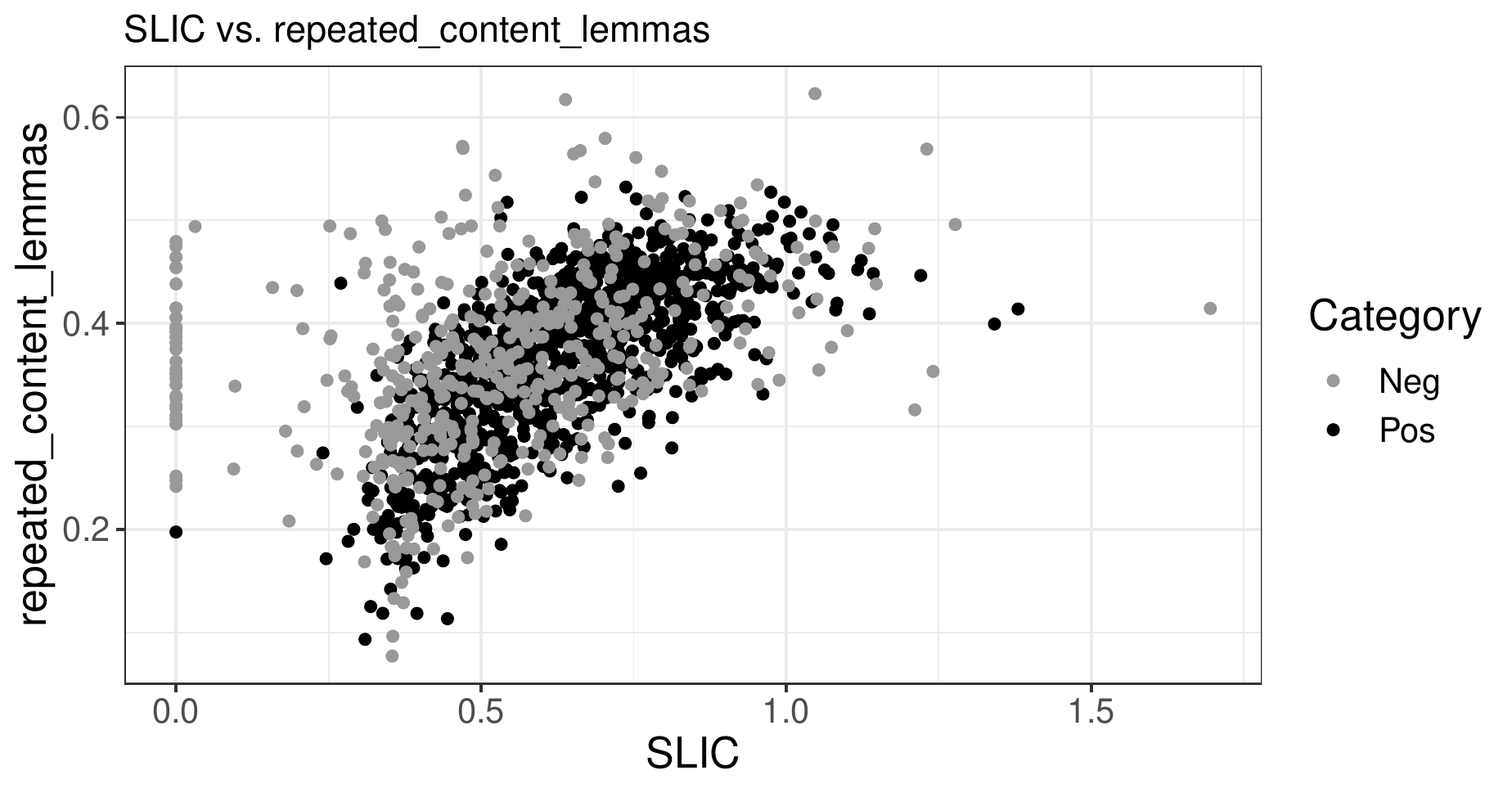}
    \caption{Correlation plot of SLIC vs. \texttt{repeated\_content\_lemmas}. Pearson Correlation Coefficient $\approx 0.544$.}
    \label{fig:slic-vs-repeated-content}
    \end{subfigure}%
    \hspace{\fill}
    \begin{subfigure}{0.48\columnwidth}
    \centering
    \includegraphics[scale=0.4]{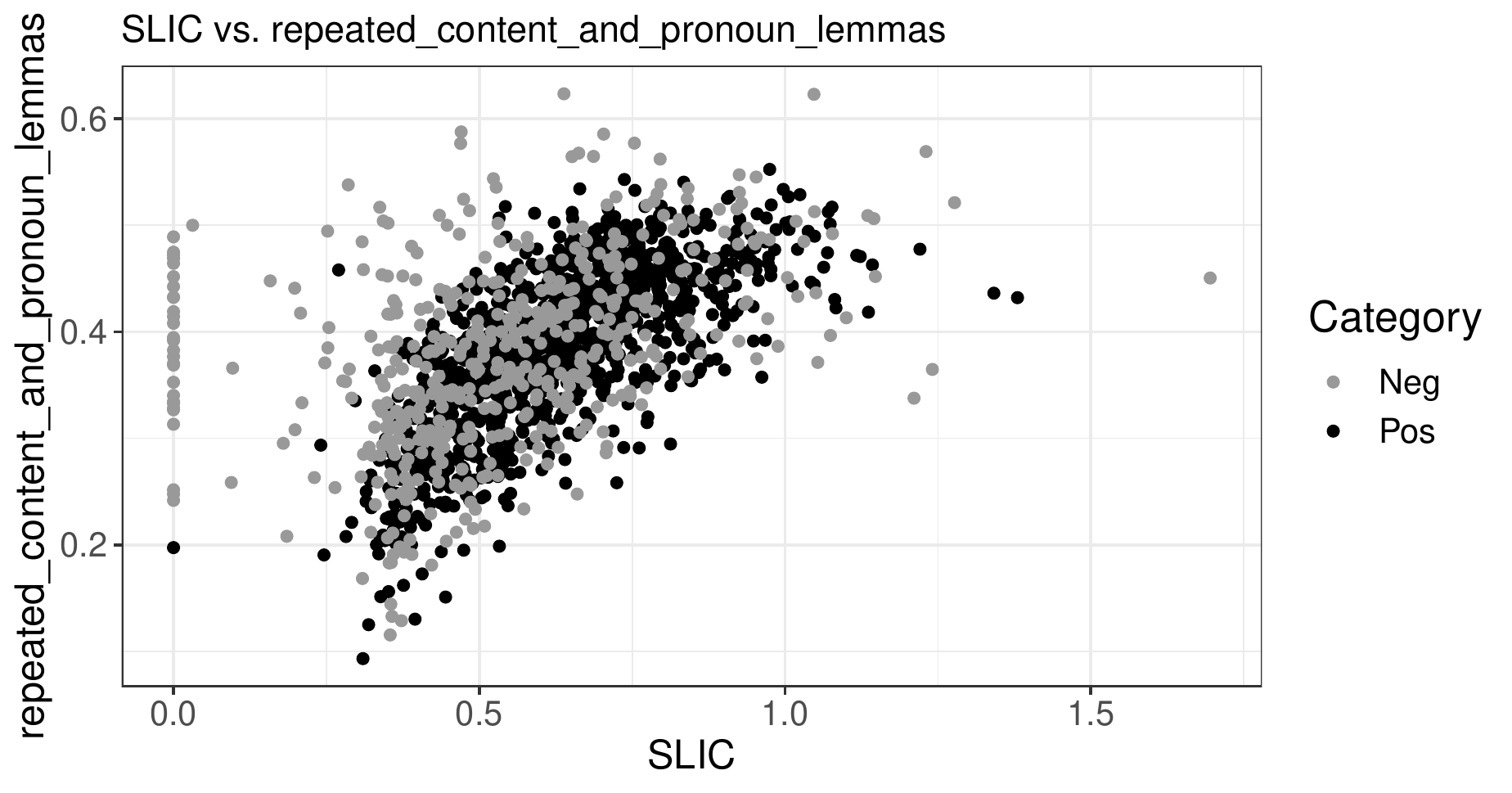}
    \caption{Correlation plot of SLIC vs. \texttt{repeated\_content\_and\_pronoun\_lemmas}. Pearson Correlation Coefficient $\approx 0.56$.}
    \label{fig:slic-vs-repeated-content-pr}
    \end{subfigure}
    
    \caption{Scatter plots showing correlation of select TAACO cohesion indices with SLIC scores for sections in `Pos' and `Neg' category documents.}
    \label{fig:slic-vs-taaco}
\end{figure}

We present correlation analysis of SLIC scores against these indices in Figure \ref{fig:slic-vs-taaco}. As expected, SLIC shows negative correlation with \texttt{lemma\_ttr} (Figure \ref{fig:slic-vs-lemma-ttr}) and \texttt{content\_ttr} (Figure \ref{fig:slic-vs-content-ttr}). Pearson correlation coefficients for \texttt{lemma\_ttr} and \texttt{content\_ttr} against SLIC are $-0.637$ and $-0.63$, respectively, which show a high degree of negative correlation. This indicates that a cohesive text with a high SLIC score exhibit lower TTR-based scores, implying that tokens are repeated in the text. On a similar note, we observe that SLIC shows positive correlation with \texttt{repeated\_content\_lemmas} and \texttt{repeated\_content\_and\_pronoun\_lemmas} indices. Pearson correlation coefficient for these two repeated content based indices are $0.544$ and $0.56$, respectively. This indicates that a text with a high SLIC score exhibit a fairly high repetition of content lemmas, which indicate lexical cohesion within the text.

\subsubsection{Distribution of SLIC Scores}
\begin{figure}[!htbp]
    \centering
    \includegraphics[scale=0.8]{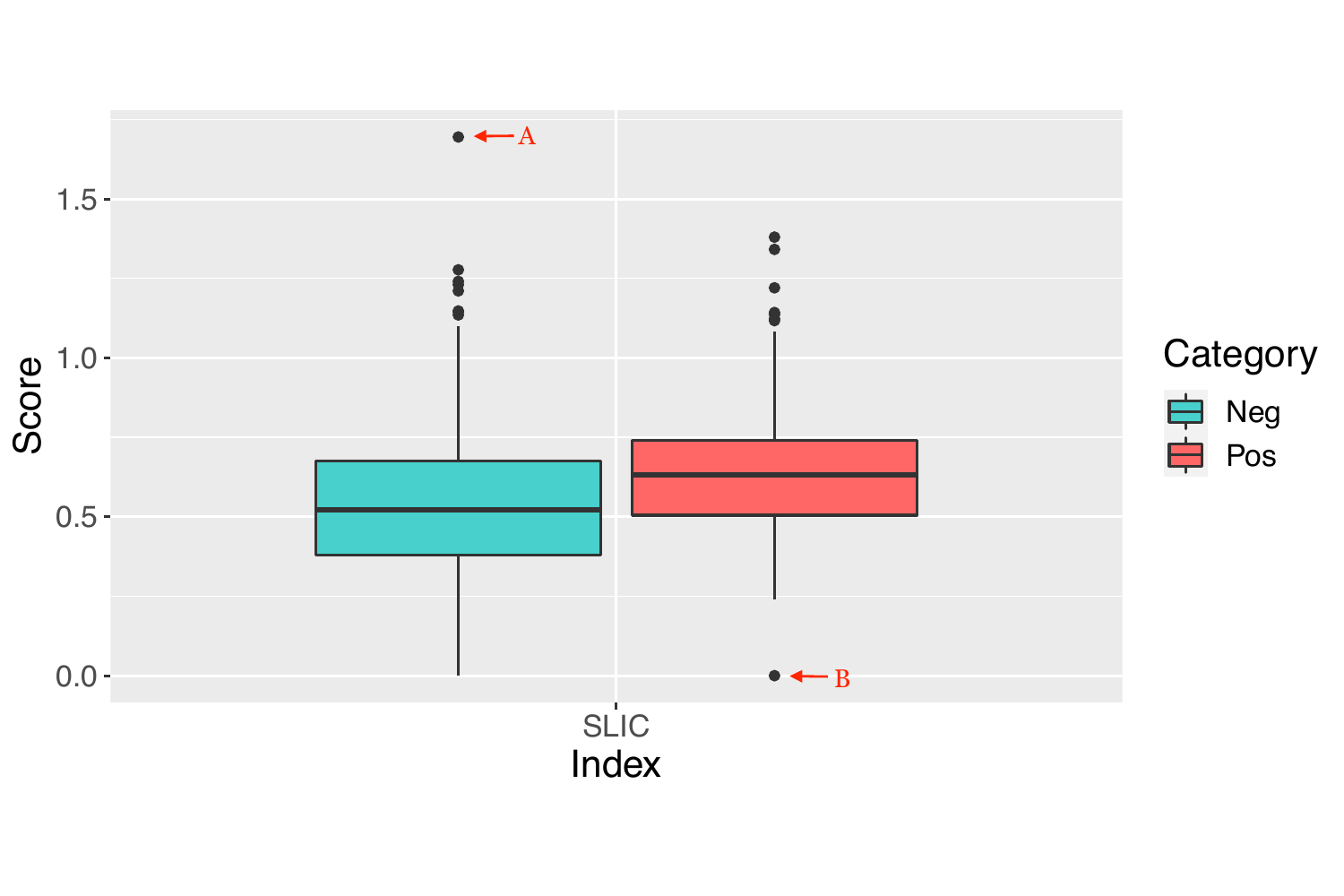}
    \vspace{-0.5cm}
    \caption{Boxplot of SLIC score for the two categories of data.}
    \label{fig:slic-boxplot}
\end{figure}

We empirically observe that the SLIC scores for the `Pos' and `Neg' category section texts in our dataset range between $[0, 1.695078]$, with a median $= 0.6019773$. We show the boxplot of the scores in Figure \ref{fig:slic-boxplot} for the two categories of data.

It is evident that the distribution of SLIC scores for the two categories of texts is not overly distinct, with a fair amount of overlapping for higher SLIC score region. Majority of `Pos' category section texts exhibit SLIC scores higher than the median of SLIC scores for `Neg' category section texts. This observation supports our hypothesis that `Pos' category sections in-general exhibit higher cohesion than `Neg' category sections. However, we reiterate that all sections from articles published in prestigious venues do not maintain the same level of sophistication in writing, and likewise, sections from articles published in predatory can be cohesive and well-written. 

There exist two examples in the boxplot, one each for `Pos' and `Neg' categories, which exhibit extreme anomalous behavior. A section text from `Neg' category exhibit highest SLIC score (outlier point \textit{A} in Figure \ref{fig:slic-boxplot}), which surpasses all `Pos' category texts. This section is from document id `doc86', section 2. The network representation for this section is near-complete with high variation in the edge weights, making the standard deviation higher than the mean, which results in a SLIC score greater than 1. Upon closer inspection, we observed that this section has only four key-entities, out of which two key-entities are repeated multiple times in the text. The other pairs are also repeated, but the frequency is significantly less. Since SLIC score is influenced by the edge weights in the network, multiple repetition of a pair of key-entities introduce high variation to the sequence of edge weight (i.e. $\sigma_W > \mu_W$ in eq. \ref{eq:slic}), which in turn increases the SLIC score. The repetition of the key-entities do signify lexical cohesion, signalling that the text in question is indeed cohesive.

Contrastingly, a section in `Pos' category also exhibits anomalous behavior, where the $SLIC$ score is 0 (outlier point \textit{B} in Figure \ref{fig:slic-boxplot}). This section is the conclusion section (section 6) from document id `478'. Upon deeper investigation, we observe that this section results into a network with two components, as multiple consecutive sentence pairs exhibit BERT NSP scores below the data-driven threshold. This results in key-entities being weakly connected to each other. Moreover, the key-entities are not repeated throughout the section which leads to the co-occurrence weights to be 1 for all edges. Consequently, the standard deviation for the sequence of edge weights becomes 0, which results in the SLIC score being 0 as well.

\subsection{Empirical Validation of Document-level Cohesion Metrics}
\label{subsec:K4-APL-res}
\begin{figure}[!hbtp]
    \centering
    \includegraphics[scale = 0.7]{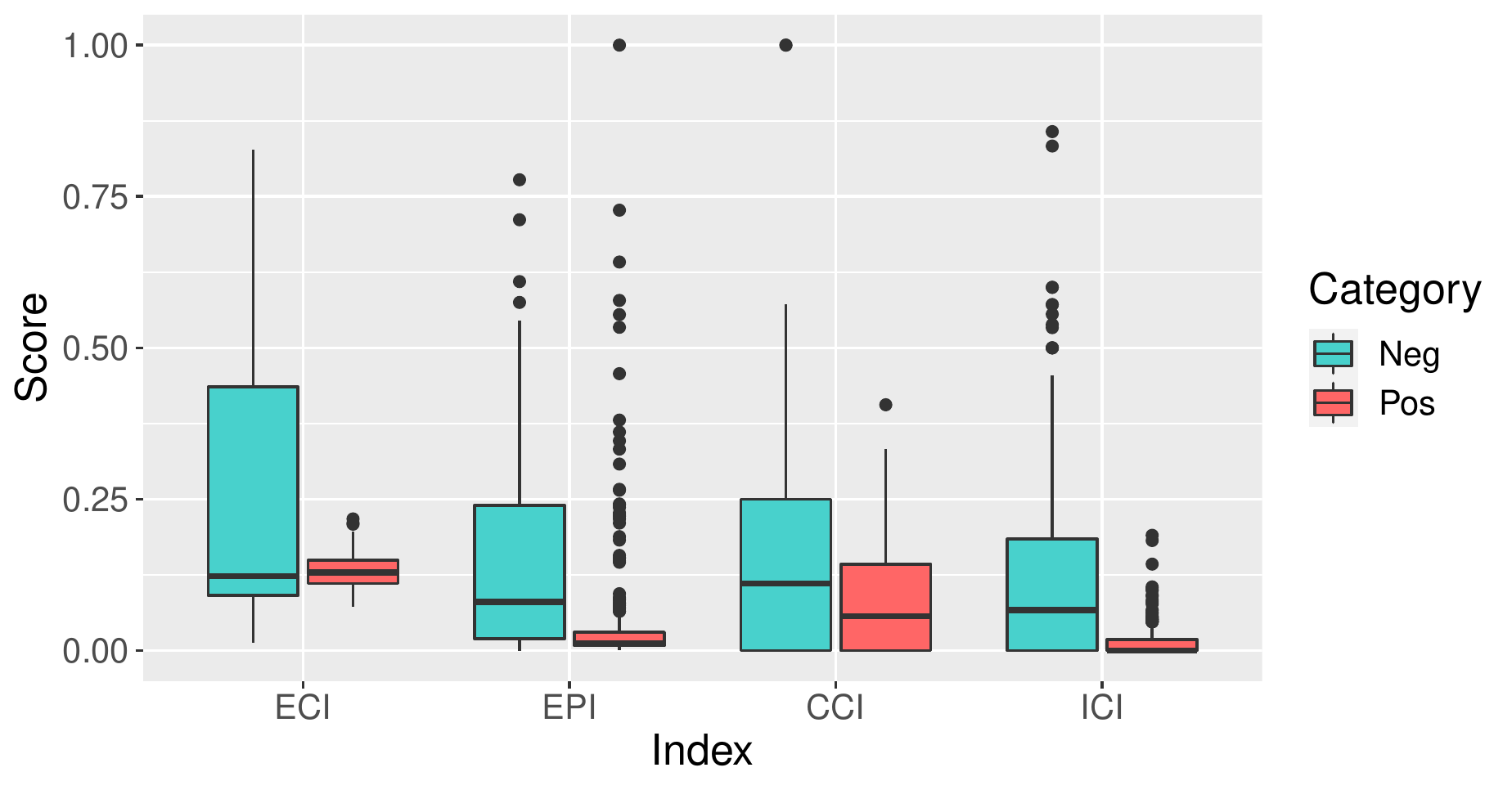}
    \caption{Boxplot of the four document-level cohesion indices based on the document category. For all four metrics, higher values indicate higher cohesion gap.}
    \label{fig:boxplot-doc-level}
\end{figure}

We establish efficacy of the four proposed metrics - $ECI$, $EPI$, ${CCI}$, and $ICI$ - for document-level cohesion by inspecting their propensity to discriminate between `Pos' and `Neg' category texts (Section \ref{sec:data}). We compute the four metrics for all documents in the `Pos' and `Neg' category and study the distribution of scores in boxplots (Figure \ref{fig:boxplot-doc-level}). Scant overlap between the $ECI$, $EPI$, and $ICI$ scores of the two categories indicates that the extracted signals have potential to discriminate between two categories of scholarly articles. Since the average shortest path length (APL) of a layer network $L_i$ can take any arbitrary value, we apply min-max normalization to scale the values of $EPI$ within the range $[0,1]$. Distribution of scores in the figure shows that ${ECI}$ and ${ICI}$ attributes are effective discriminators, whereas ${CCI}$ is least discriminating. We reiterate that both `Pos' and `Neg' category contain documents that exhibit anomalous scores for the document-level cohesion indices. This indicates that some documents in `Pos` category lack lexical cohesion and require improvement in writing, which are evident as outliers in `Pos' category. Outliers in `Neg' category are indeed documents with weak writing that exhibit cohesion gaps.

It is reasonable to conclude that the proposed cohesion metrics for the two classes of articles, i.e, published in prestigious and predatory venues, exhibit distinct network properties. Articles in prestigious venues (in this case, ICLR) maintain a high level of writing quality which is evident in the form of coherence and cohesion in the text. Setting aside factors like problem significance, technical correctness, and novelty, the writing quality of articles published in prestigious venues is enforced by a rigorous peer-review process that maintains a high expectation. On the other hand, articles published in predatory venues exhibit a comparatively weak quality of writing evident from low (and often, lack of) coherence and cohesion in the text. The weak writing quality often gets disregarded due to more relaxed expectations in the review process.

\subsubsection{Discriminating Ability of the Document-level Cohesion Indices} 
To assess the discriminating ability of the four indices, we perform single-attribute classification on the four individual properties using a decision tree based classifier - C4.5 algorithm \citep{Quinlan1993}. We adopt the J48 implementation of C4.5 algorithm available on WEKA platform \citep{eibe2016weka} and approach the problem as a binary classification task. Table \ref{tab:classification-global-coh} presents the 10-fold cross-validation of the J48 algorithm on the four individual properties as well as together.

\begin{table*}[!htbp]
    \centering
    \begin{tabular}{c|c|c|c|c|c|c|c}
    \hline
        \multirow{2}{*}{\textbf{Attribute}} & \multirow{2}{*}{\textbf{Accuracy}} & \multicolumn{2}{c}{\textbf{Precision}} & \multicolumn{2}{c}{\textbf{Recall}} & \multicolumn{2}{c}{\textbf{F1-score}}\\ \cline{3-8}
        & & $\mathbf{Pos}$ & $\mathbf{Neg}$ & $\mathbf{Pos}$ & $\mathbf{Neg}$ & $\mathbf{Pos}$ & $\mathbf{Neg}$ \\ \hline
        $\mathbf{ECI}$ & \textbf{80.07} & 0.797 & \textbf{0.811} & 0.918 & 0.600 & \textbf{0.853} & \textbf{0.690} \\
        $\mathbf{EPI}$ & 69.00 & \textbf{0.827} & 0.558 & 0.643 & \textbf{0.770} & 0.724 & 0.647 \\
        $\mathbf{CCI}$ & 68.27 & 0.672 & 0.792 & \textbf{0.971} & 0.190 & 0.794 & 0.306 \\
        $\mathbf{ICI}$ & 73.43 & 0.728 & 0.759 & 0.924 & 0.410 & 0.814 & 0.532 \\  \hline
        \textbf{All} & 81.55 & 0.827 & 0.791 & 0.895 & 0.680 & 0.86 & 0.731 \\ \hline
    \end{tabular}
    \caption{10-fold CV results of the four metrics (attributes), both individually and together, using a J48 classifier. Bold values indicate best performance among the four single-attribute classifiers in the corresponding column.}
    \label{tab:classification-global-coh}
\end{table*}

The results assert that ${ECI}$ is the most discriminating and ${CCI}$ is the least discriminating attribute among the four. We observe that ${ECI}$, ${{ICI}}$, and $CCI$ report a higher true-positive rate for the positive category (label 1 = `Pos'), whereas ${EPI}$ report a higher true-positive rate for the negative category (label 0 = `Neg'). This means that ${ECI}$, ${{ICI}}$, and $CCI$ are able to correctly predict the labels for most of the positive category documents. On the other hand, ${EPI}$ is able to correctly predict the labels for most of the negative category documents. Specifically, $CCI$ as a feature is biased towards the `Pos' category, where the model assigns label 1 (= `Pos') to majority of documents irrespective of the ground truth. On the other hand, ${EPI}$ exhibits a contrasting behavior by correctly predicting the class labels of majority of the negative class documents with a penalty of incorrectly labeling more than one-third of the positive class documents. This contrasting behavior of the four attributes pays off, as the combined behavior of all the metrics (row corresponding to `All' attributes in Table \ref{tab:classification-global-coh}) captures best of the both worlds and is able to achieve satisfactory results for both categories.

It is noteworthy that our objective is \textit{not} to classify scholarly documents into binaries of ``well" or ``poorly" written documents using the publication venue (prestigious vs. predatory venues) as ground truth categories. Therefore, we do not attempt to optimize for classifier accuracy. Instead, we aim to analyze both classes of documents for cues that indicate the quality of writing in terms of cohesion. 

\section{CHIAA Framework}
\label{sec:author_suggestion}
Multilayer network representation of scientific scholarly texts lays the groundwork for the analytical framework \textit{CHeck It Again, Author} (CHIAA), which facilitates objective assessment of overall writing quality and effective scrutiny of cohesion gaps in the text. Admitting due emphasis to explainable decisions by machines, \textit{CHIAA} provides clear and precise leads to the author for enhancing the writing quality by improving lexical cohesion in the text. Specifically, we identify the following features as salient cues.

\begin{enumerate}
    \item \textit{Regions of low  cohesion at section level:} The author of the text is advised to review sections in the text that correspond to a comparatively low $SLIC$ score or with more than one component. Armed with relevant information, the author needs to scrutinize the corresponding pairs of sentences with low BERT NSP scores and improve the writing quality. For example, cohesion gap is evident in Figure \ref{Fig:writing-quality}, which shows a disconnected network (Figure \ref{subfig:two-comp}) for the corresponding text (Figure \ref{subfig:bad-text}). Consequently, the $SLIC$ score for this text is computed as $0.3543$, which is closer to the lower bound. In this particular case, the author may improve the lexical cohesion by inspecting the highlighted portion in Figure \ref{subfig:bad-text}, which corresponds to sentence pair (4,5) (Figure \ref{subfig:coh-plot}).
    
    \item \textit{Weakly connected entities in layers:} Sections that contain weakly connected key-entities due to fewer intralayer edges in $\mathcal {M}$ exhibit semantic gap among key-entities present in the section. In CHIAA framework, such sections exhibit high $ECI$ and $EPI$ values and are therefore flagged as sections with cohesion gaps. The author is supplied relevant information to look into the affected concepts (constituent entities) and review writing. For example, document id `doc16' (used as an example in Figures \ref{Fig:writing-quality}, \ref{fig:community-detection-example}, \ref{fig:community-metagraph}) has $ECI = 0.1743929$ and $EPI =  6.553249$, where we observe a relatively low $ECI$ with a high value for $EPI$. The high value of $EPI$ denotes longer paths between the pairs of nodes in network $\mathcal {M}$, indicating weak semantic connections and cohesion among the key-entities. Since $EPI$ is affected by the deviation of $APL_i$ from the ideal value of $\log(n_i)$, the prescribed suggestion by CHIAA is to target and improve writing in the sections with the lowest negative deviation $(\log(n_i)-APL_i)$. In this particular case, the author is suggested to review writing in sections 4 and 6, where the deviation is approximately $-9$ and $-14$, respectively. Accordingly, if linked appropriately, the entities in section 4 (and section 6) could improve lexical cohesion in the section and overall document.
    
    \item \textit{Weak cohesion of concepts in document:} Weak cohesion among concepts within and across sections is captured by $ICI$ and $CCI$ metrics, where values farther from 0 indicate cohesion gaps. The absence of interconnecting metaedges in $\mathcal {\widehat M}$ indicates that two concepts in adjacent sections bear weak or no semantic connection. The author may review to strengthen the linkage between the marked concepts in two sections. For example, the metagraph shown in Figure \ref{fig:community-metagraph} has $ICI = 0.2173913$ (relatively higher than most documents of the `Pos' category), where we have one isolated concept each in layers 2 and 6, and three (i.e., all) isolated concepts in layer 4. In these three layers (sections), the problematic concepts bear weak or no semantic connection to the other concepts within that section. Appropriately improving the writing around the concepts discussed in layers 2, 4, and 6 could improve cohesion in those sections. On the other hand, the same metagraph exhibit $CCI \approx 0.032$, implying there is a nominal change in the count of complete subgraphs in $\mathcal {\widehat M}$ before and after edge pruning. This is because, three (out of 56) interlayer metaedges  were pruned from $\mathcal {\widehat M}$ in Figure \ref{fig:community-metagraph} (metaedges between nodes (4.2, 5.2), (4.3, 5.1), and (4.3, 5.2)), in addition to the eight intralayer metaedges (out of 23) in sections 2, 4, and 6. This indicates that the cohesion among concepts across sections is fairly consistent, and improving the interlayer connectivity between concepts discussed in sections 4 and 5 could further enhance lexical cohesion across sections in the document. Cohesion among concepts within sections requires improvement in sections 2, 4, and 6, as indicated by both $ICI$ and $CCI$ metrics.
\end{enumerate}

\textit{CHIAA} framework for qualitative assessment of cohesion based on the indices is transparent. Scores reveal weakness of each aspect and offer precise and concise explanations, which are readily accessible to the author. Recommendations for strengthening regions in the text for each aspect are generated to help the author in improving the text quality in terms of cohesion. 

\subsection{CHIAA Analysis of our Manuscript}
We perform CHIAA analysis of our manuscript at the section and document level. We extract the section text from each section, including the abstract and conclusion sections, and perform standard text cleaning to remove section and subsection headings, figures and tables along with captions, and mathematical equations. We exclude acknowledgment, appendix, references, and this subsection from our analysis. We apply section-level and document-level analysis to all the sections and report our observations below.

In the section level analysis, we observe that the SLIC scores for the sections range from $0.4365397$ to $1.1160087$, where values closer to 0 indicate cohesion gaps in the section texts. In our case, the abstract section exhibit the lowest SLIC score ($< 0.5$). The rest of the sections exhibit SLIC scores greater than $0.5$. Since the abstract presents a concise summary of the work presented in the manuscript, it is expected to contain some key-entities which may not have a strong semantic relationship among themselves. In the document level analysis, this manuscript scores $ECI = 0.14727895571731$, $EPI = 0.114096083780721$, $CCI = 0.1$, and $ICI = 0$. We observe that these scores are closer to 0, implying that the manuscript exhibits lexical cohesion at the document level, both within and across sections. 

\section{Discussion}
\label{sec:implications}
A well-written text is expected to exhibit a lexically cohesive discourse, without notable cohesion gaps. Cohesion gaps in a scientific scholarly text negatively affect the readers' ability to form a coherent narrative, impeding comprehension of the authors' intended meaning conveyed in the text. 

Lexical cohesion among entities in the text is determined by how well the discourse entities are knit together throughout the text. Existing cohesion analysis tools, such as Coh-Metrix \citep{graesser2004coh, graesser2014coh} and TAACO \citep{crossley2016tool,crossley2019tool}, provide the user with numerous qualitative metrics (cohesion indices) to gauge the extent of cohesion in the discourse by analyzing various syntactic and semantic aspects of the text. However, the sheer number of the cohesion indices computed by the tools and difficulty in their interpretation present a barrier in its popularity among users outside the \textit{linguistic} research community. Moreover, the absence of exclusive thresholds for the indices limits the direct practical utility of these indices for a general user. Furthermore, both these tools do not assist the user with pointers to regions of text that exhibit cohesion gaps, thereby overlooking the prescriptive applicability.

Multilayer representation of scientific scholarly text is a detailed non-linear map of the lowest level semantic units (key-entities), with intricate details of their relationship encoded at the higher level units (sections and the whole document). Modeling each section as network enables scrutiny of semantic connections between the entities referred to in the section. Condensing the layer $L_i$ to $\hat L_i$ reveals the degree of semantic relatedness between the latent concepts encapsulating the entities within the section, as proposed in this work. The strength of interlayer edges divulges how well the latent concepts are intertwined across sections, which determines the cognitive load imposed by the text. In  the interest of simplicity, we omit intermediate semantic units, such as paragraphs, subsections, etc., in the current study. We believe that incorporating intermediate-level semantic units would enable a more detailed and fine-grained analysis of the discourse. Identifying domain-specific hierarchies, varying the types of nodes (words, phrases, sentences, part-of-speech tags, etc.) and types of relationships (similarity, synonymy, hyponymy, collocation, etc.) opens up avenues for a multitude of analyses of news articles, reports, scholarly texts, legal documents etc. The inherent potential is a significant theoretical implication of this innovative representation of text.

The analytical framework, \textit{CHIAA}, which provides prescriptive suggestions to the user based on computed metrics of text cohesion, is a notable outcome with significant practical utility. The section-level and document-level cohesion metrics are transparent and easy to understand and provide the user with intuitive guidelines for assessing the writing quality of the text. CHIAA framework provides the user with practical suggestions to improve the writing by locating regions in the text with cohesion gaps. Our work sets the direction for designing and developing more and better tools to assist students, research scholars, and authors from countries with English as a second or foreign language in composing high-quality texts.



\section{Conclusion}
\label{sec:conclusion}
We propose a novel representation of scientific scholarly documents as multilayer network (MLN) and present an analytical framework \textit{CHIAA} for qualitative assessment of writing quality in terms of lexical cohesion. Each section in the text is represented as a layer in the multilayer network, and is modeled as an undirected weighted network with key-entities as nodes and co-occurrence relations as edges. We exploit pertinent signals distilled from individual layer networks that proxy for cohesion in the section text. Furthermore, the nodes in two consecutive layers are interconnected with each other through semantic similarity. The multilayer network representation is further condensed to uncover latent semantic relations among concepts encapsulating the key-entities discussed in the text. We derive network-based properties from the condensed multilayer network to quantify document-level cohesion in the text. The CHIAA framework provides prescriptive suggestions for improving writing quality based on the section-level and document-level cohesion analysis.

We present a proof-of-concept for our proposed approach using a mixed dataset of a publicly available collection of articles from prestigious venues and a curated set of scholarly articles published at predatory venues. Preliminary investigation for establishing the efficacy of the MLN representation model and \textit{CHIAA} framework has been insightful. 

\section*{Acknowledgement}
\label{sec:ack}
This work is supported by Department of Science and Technology, Govt. of India, grant MTR/2019/000604.

\newpage
\appendix
\section{Example of Situations that Induce Cohesion}
\label{app:example-cohesion}

\citet{halliday1976cohesion} identifies five situations that induce cohesion - reference, substitution, ellipsis, conjunction, and lexical cohesion. We briefly describe these situations below.
\begin{enumerate}
\item \textit{Reference} involves usage of pronouns (he, she, they, etc.), comparatives (e.g., the more \textit{x} than ), etc. to refer to some entity. For example, in the sentence pair `\textbf{John} went to the store. \textbf{He} bought a packet of milk.', the entity \textit{John} is replaced by \textit{he} in the second sentence.

\item \textit{Substitution} involves replacing an entity in a text with another. For example, in the sentence `My \textbf{laptop} broke down. So I bought a new \textbf{one}.', the word \textit{laptop} is replaced by \textit{one} in the second sentence.

\item \textit{Ellipsis} is a type of cohesion where an entity is omitted or replaced by nothing. For example, the following exchange - `Are you hungry? - Yes, I am.' - involves omitting the word \textit{hungry} from the response.

\item \textit{Conjunction} involves using connectors to connect sentences. These connectors include words and phrases like \textit{and, for, but, so, furthermore, accordingly, therefore, as a result, etc.}. The sentences - `I was hungry. \textbf{So} I ate an apple.' - is an example of such type of cohesion.

\item \textit{Lexical cohesion} in text signifies the semantic relationship among words, which can be captured as recurrence of the words and their synonyms. For example, the sentences - `I have a \textbf{cat}. The \textbf{cat} is sitting on the mat.' - exhibit lexical cohesion by repetition of the word \textit{cat}. In this study, we focus on assessing lexical cohesion in textual discourse.
\end{enumerate}

\newpage

\bibliographystyle{model5-names}

\bibliography{cohesion}

\end{document}